\documentclass[lettersize,journal]{IEEEtran}

\IEEEoverridecommandlockouts

\usepackage{amsmath,amssymb,amsfonts}
\usepackage{array}
\usepackage[caption=false,font=normalsize,labelfont=sf,textfont=sf]{subfig}
\usepackage{textcomp}
\usepackage{stfloats}
\usepackage{url}
\usepackage{verbatim}
\usepackage{graphicx}
\usepackage{cite}
\usepackage{xspace}
\newcommand \name{Flow Motion Policy\xspace}

\usepackage[hidelinks]{hyperref}
\hypersetup{
   linkcolor=blue,
   breaklinks=true,   
   colorlinks=true,   
   citecolor=blue,
   urlcolor=blue
}
\usepackage{xcolor}
\usepackage{siunitx,booktabs}
\usepackage{multirow}
\usepackage{caption}
\usepackage{hhline}
\usepackage{comment}

\usepackage{censor}
\def\censorcolor{gray!50} \let\svcensorrule\censorrule \renewcommand\censorrule[1]{ \textcolor{\censorcolor}{\svcensorrule{#1}}}

\usepackage[ruled,vlined,linesnumbered]{algorithm2e}

\SetCommentSty{mycommfont}
\SetKwComment{Comment}{// }{}
\let\oldnl\nl%
\newcommand{\nonl}{\renewcommand{\nl}{\let\nl\oldnl}}%

\usepackage[table]{xcolor}
\usepackage[switch]{lineno}
\usepackage[most]{tcolorbox}

\begin{document}

\title{Flow Motion Policy: \\
\LARGE{Manipulator Motion Planning with Flow Matching Models}}

\author{Davood Soleymanzadeh$^{1}$, Xiao Liang$^{2,*}$, and Minghui Zheng$^{1,*}$
\thanks{$^{1}$Davood Soleymanzadeh and Minghui Zheng are with the J. Mike Walker '66 Department of Mechanical Engineering, Texas A\&M University, College Station, TX 77843, USA (\tt\footnotesize e-mail: davoodso@tamu.edu; mhzheng@tamu.edu).}
\thanks{$^{2}$Xiao Liang is with the Zachry Department of Civil and Environmental
Engineering, Texas A\&M University, College Station, TX 77843 USA (\tt\footnotesize e-mail:
xliang@tamu.edu).}
\thanks{$^*$ Corresponding Authors.}
\thanks{This work was supported by the USA National Science Foundation under Grant No. 2527316 and No. 2422826. Portions of this research were conducted with the advanced computing resources provided by Texas A\&M High Performance Research Computing.}
}

\makeatletter
\let\@oldmaketitle\@maketitle%
\renewcommand{\@maketitle}{\@oldmaketitle%
\setcounter{figure}{0}
\vspace{12pt}
\centering
\begin{center}
\includegraphics[width=1.0\linewidth]{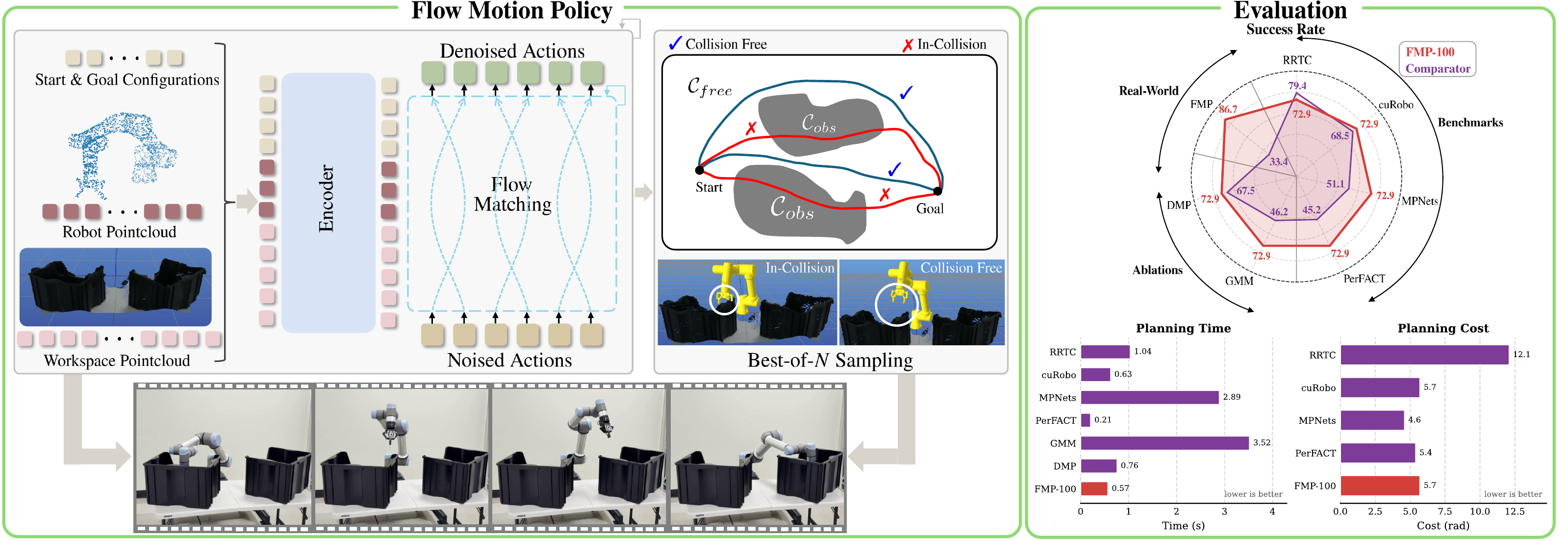}
  \captionof{figure}{\small \textbf{Method Overview.} \name (FMP \& FMP-100) is a point cloud–conditioned flow-based motion policy that enables best-of-$N$ sampling (FMP-100) for motion planning in complex environments. The framework leverages the generative formulation of flow matching to capture the inherent distribution of motion planning datasets and supports efficient inference-time best-of-$N$ sampling. Compared to benchmark planners such as RRT-Connect (RRTC)~\cite{kuffner2000rrt}, cuRobo~\cite{sundaralingam2023curobo}, and MPNets~\cite{qureshi2020motion} and end-to-end deterministic neural motion planners such as PerFACT~\cite{soleymanzadeh2025perfact}, \name achieves higher planning efficiency. Furthermore, the inference-time best-of-N sampling with \name is more efficient than using Gaussian Mixture Model (GMM)- or diffusion (DMP)-based head architectures within the same policy architecture. $\mathcal{C}_{free}$ and $\mathcal{C}_{obs}$ are free configuration space and obstacle configuration space, respectively.}
  \label{fig:openning}
  \end{center}
  \vspace{-0.2in}
  }
\makeatother
\maketitle
\thispagestyle{empty}
\pagestyle{empty}
\begin{abstract}
Open-loop end-to-end neural motion planners have recently been proposed to improve motion planning for robotic manipulators. These methods enable planning directly from sensor observations without relying on a privileged collision checker during motion planning. However, existing planners produce a single path for a given planning problem and cannot exploit their open-loop nature to propose multiple motion plans. To address this limitation, we introduce Flow Motion Policy, an open-loop neural motion planner that uses flow matching to generate a batch of motion plan proposals by learning a distribution over motion plans conditioned on the planning observation. At inference time, it samples multiple candidate motion plans to enable efficient best-of-$N$ inference while avoiding iterative collision checking during planning. We benchmark the Flow Motion Policy against representative sampling-based, optimization-based and neural motion planning methods. Evaluation results demonstrate that Flow Motion Policy improves planning success and efficiency, highlighting the effectiveness of stochastic generative policies for end-to-end motion planning and best-of-$N$ sampling. Project website: \href{https://davoodsz.github.io/FlowMotionPolicy.github.io/}{https://davoodsz.github.io/FlowMotionPolicy.github.io/}
\end{abstract}

\section{Introduction}
Motion planning plays an important role in robotic manipulation systems, operating alongside high-level control policies to ensure safe and feasible actions. The objective is to compute a valid, collision-free path that connects an initial configuration to a desired goal configuration~\cite{noroozi2023conventional, soleymanzadeh2026towards}. Achieving fast and reliable motion planning for manipulators in unstructured environments remains a challenging problem. Decades of research have established many algorithms to achieve this, among which probabilistically complete sampling-based~\cite{lavalle1998rapidly} and trajectory optimization~\cite{zucker2013chomp} algorithms have been widely used for robotic manipulators.

Sampling-based planners construct a tree in the configuration space to connect the start and goal configurations by iteratively generating random samples within the configuration space, steering towards the newly sampled configurations, and collision-checking the resulting connections~\cite{lavalle2001rapidly, karaman2011sampling}. Due to their stochastic sampling process, these planners may generate different plans for the same planning problem across multiple runs. However, these planners rely on a privileged collision checker during planning, and collision checking is the main bottleneck in motion planning, accounting for up to 90\% of the computation time~\cite{das2020forward}, which makes parallelism and best-of-$N$ sampling challenging~\cite{huang2025prrtc, thomason2024motions}. Although recent advanced variants~\cite{zhang2024flexible, gammell2020batch} introduce heuristic-based informed samplers to bias the sampling process towards the goal to reduce the number of collision queries, such heuristics are challenging to design for high-dimensional configuration spaces. 

Optimization-based methods~\cite{zucker2013chomp, kalakrishnan2011stomp} iteratively refine an initial trajectory, often a straight-line interpolation between the start and goal configurations, using numerical optimization techniques. These methods enforce constraints, including obstacle avoidance and trajectory smoothness, during optimization. Although trajectory optimization approaches are amenable to parallelization and can benefit from best-of-$N$ initialization strategies~\cite{sundaralingam2023curobo}, they still require a privileged collision checker to provide both collision status and gradient information throughout the optimization process.

Recently, deep learning methods have been leveraged to address the limitations of sampling-based and optimization-based planners. In sampling-based planning, neural networks have been proposed as neural-informed samplers~\cite{qureshi2020motion,soleymanzadeh2025simpnet, johnson2023learning, soleymanzadeh2026gaide} to bias tree expansion towards the goal, reducing the number of collision checks. Neural networks have also been used as implicit collision checkers~\cite{kim2023pairwisenet, song2024graph}, enabling faster collision queries. However, these approaches still require collision querying during planning, which limits their efficiency in best-of-$N$ sampling. For optimization-based methods, learning-based methods have been leveraged to generate high-quality initial trajectories~\cite{carvalho2025motion, yan2025m, nguyen2025flowmp}, which can help to avoid local minima and accelerate convergence. Nevertheless, these methods also depend on collision checkers to provide collision status and gradient information during optimization. 

Moreover, another line of work leverages deep learning for open-loop, end-to-end motion planning~\cite{soleymanzadeh2025perfact, dalal2024neural, yang2025deep}, eliminating the need for a privileged collision checker during planning. These methods learn a direct mapping from the planning problem information (e.g., start configuration, goal configuration, and environment observations) to a feasible path. However, most existing approaches are deterministic policies that produce a single path for a given planning problem. As a result, they cannot exploit best-of-$N$ sampling at inference time.

We present \name, an end-to-end neural motion planner that leverages the stochastic generative formulation of flow matching frameworks~\cite{lipman2022flow} to enable best-of-N sampling during inference, as demonstrated in Figure~\ref{fig:openning}. \name generates multiple candidate paths and applies a light-weight best-of-$N$ sampling to select the first collision-free solution for execution. This design allows \name to operate both as a direct end-to-end planning policy (without planning), and as a planner enhanced by best-of-$N$ sampling at inference time.

Our main contributions are:
\begin{itemize}
    \item We use flow matching to develop a learned stochastic motion planning policy for robotic manipulators. The policy can directly generate a path or generate multiple candidate paths for best-of-$N$ selection, where the first collision-free path is selected for execution.
    \item We evaluate \name on held-out evaluation tasks, and demonstrate its effectiveness compared to state-of-the-art motion planners.
\end{itemize}

The rest of the paper is organized as follows. Section~\ref{sec: relatedwork} presents an overview of prior work in motion planning for robotic manipulators. Section~\ref{sec: graphtransformer} describes the proposed \name framework. Section~\ref{sec: results} reports evaluation results and compares \name with benchmark motion planning methods. Finally, Section~\ref{sec: conclusion} concludes the paper.
\section{Related Works} \label{sec: relatedwork}
We survey related work on motion planning for robotic manipulators, and outline the main differences of the proposed framework and the state of the art.

\begin{figure*}[t]
    \includegraphics[width=0.9\linewidth]{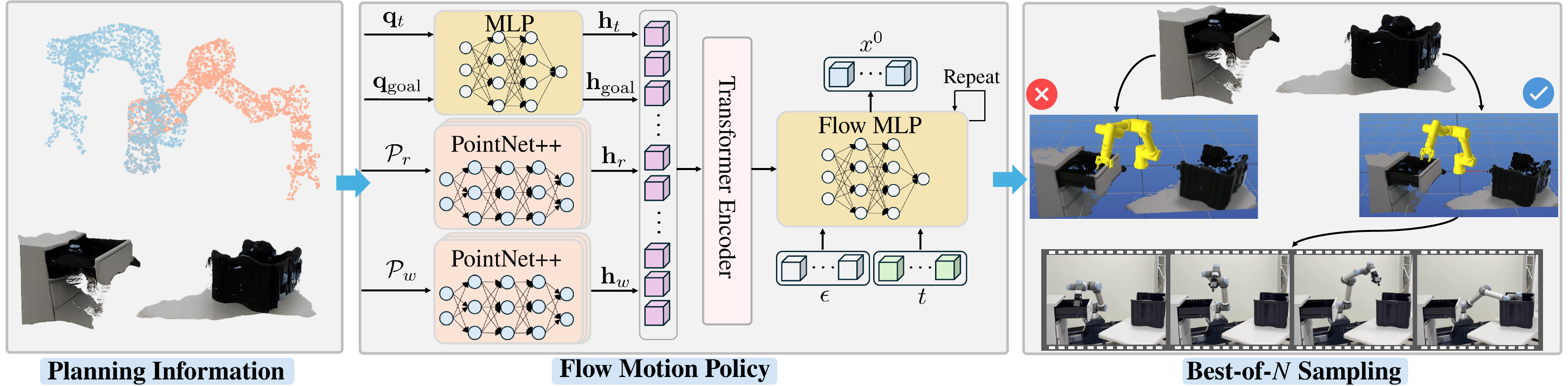}
    \centering
    \captionof{figure}{\small \textbf{\name architecture.} This framework leverages current time-step $t$ planning information (i.e., planning scene ($\mathcal{P}_w$) point cloud, robot ($\mathcal{P}_r$) point cloud, robotic manipulator configuration ($\mathbf{q}_t$)) together with robotic manipulator goal  configuration ($\mathbf{q}_\text{goal}$) for end-to-end open-loop motion planning in complex environments. $\mathbf{h}_t$, $\mathbf{h}_r$, and $\mathbf{h}_w$ are the current time-step $t$ planning information corresponding embeddings, respectively, and $\mathbf{h}_{\text{goal}}$ is the goal configuration embeddings. ``PointNet++'' are set abstraction layers from PointNet++ \cite{qi2017pointnet++}. $\epsilon \sim \mathcal{N}(0, I)$ is Gaussian noise, $t$ is flow time, and $x^0$ is denoised increment action sequence. \textbf{Best-of-$N$ Sampling} operates on a batch of planned motions and utilizes a lightweight inference-time optimization process to output the first collision-free path.}
    \label{fig:fmp}
\end{figure*}

\vspace{0.2cm}
\noindent
{\textbf{Motion Planning with Collision Checking.}} Over the years, numerous motion planning algorithms have been developed for robotic manipulators~\cite{soleymanzadeh2026towards}, with sampling-based~\cite{noroozi2023conventional} and optimization-based~\cite{zucker2013chomp} planners being the most widely used. 

Sampling-based planners iteratively build a search tree over the robot's configuration space, stochastically generate samples, and expanding the tree towards the planning goal~\cite{lavalle1998rapidly}. Due to their stochastic sampling process, these planners are inherently multi-modal, meaning that different runs can produce different feasible trajectories for the same planning problem. This property can, in principle, be exploited for parallel planning and best-of-$N$ sampling. However, these planners require a collision checker during planning, which is the dominant computational bottleneck in planning~\cite{huang2025prrtc, thomason2024motions}. Hand-crafted informed~\cite{zhang2024flexible, strub2020advanced, gammell2020batch} and neural informed samplers~\cite{qureshi2020motion, johnson2023learning,qureshi2020neural, soleymanzadeh2025simpnet, soleymanzadeh2026gaide} have been proposed to guide the planning towards goal, reducing the number of collision queries. However, these informed samplers still inherit the sequential nature of sampling-based planning algorithms and require a collision checker in the tree construction process, which is computationally expensive to perform best-of-$N$ sampling during inference time~\cite{soleymanzadeh2025simpnet}. In contrast, \name performs collision checking in parallel to enable efficient best-of-$N$ sampling during inference.  

Optimization-based planners start from an initial trajectory, which may be in collision, and enforce planning constraints and utilize numerical optimization to push the initial path to collision-free spaces while satisfying smoothness constraints~\cite{zucker2013chomp, kalakrishnan2011stomp, mukadam2018continuous, cohn2025non}. However, these planners depend strongly on the quality of the initial trajectory and can often converge to local minima. To mitigate this issue, deep learning methods have recently been leveraged to learn the underlying distribution of planning datasets to warm-start trajectory optimization~\cite{carvalho2025motion, yan2025m, nguyen2025flowmp}. Although trajectory optimization planners can operate on a batch of initial solutions and select one that satisfies planning constraints, they still require collision checking during planning, which increases computation time. In contrast, \name performs collision checking in parallel during inference for efficient best-of-$N$ sampling.

\vspace{0.2cm}
\noindent
\textbf{Open-loop End-to-end Neural Motion Planning.} Deep learning approaches have also been explored for end-to-end open-loop motion planning, where neural networks plan motions in an open-loop manner without relying on privileged collision checkers~\cite{soleymanzadeh2025perfact, yang2025deep, fishman2023motion, fishman2024avoid}. However, most existing approaches are deterministic motion policies and therefore do not support inference-time best-of-$N$ sampling. Neural MP~\cite{dalal2024neural} addresses this limitation by adopting a Gaussian Mixture Model (GMM) as the policy output representation to facilitate best-of-$N$ sampling during inference. Nevertheless, GMMs have limited capacity to represent the inherent distribution in motion planning problems. In contrast, \name employs flow matching~\cite{lipman2022flow} to model this inherent distribution and to support best-of-$N$ sampling during inference.

\vspace{0.2cm}
\noindent
{\textbf{Diffusion and flow matching models in robotics and motion planning.}} Diffusion and flow models have emerged as powerful deep generative models in robotics~\cite{chi2025diffusion}. Diffusion and flow-based behavior cloning can learn complex policies and have demonstrated strong performance in real-world robotic applications, spanning from task-specific settings~\cite{dasari2025ingredients} to generalist policies~\cite{bjorck2025gr00t, black2024pi_0}. Diffusion models have also been explored for motion planning, including end-to-end approaches~\cite{yan2025m} and for warm-starting trajectory optimization algorithms~\cite{carvalho2025motion, nguyen2025flowmp}. In this work, we employ a flow-based policy as the decoder head of the proposed open-loop, end-to-end neural motion planner, enabling best-of-$N$ sampling at inference time~\cite{zhou2024diffusion}.

\section{\name} \label{sec: graphtransformer}
In this section, we present \name, an end-to-end motion planner with a flow matching head that enables best-of-$N$ sampling at inference time. The overall framework is shown in Figure~\ref{fig:fmp}.

\subsection{Problem Definition} \label{subsec: def}
We consider motion planning for a manipulator with configuration $\mathbf{q}\in\mathbb{R}^n$, start configuration $\mathbf{q}_{\text{start}}$, goal configuration $\mathbf{q}_{\text{goal}}$, and environment observation $\mathcal{P}$. The goal is to generate a collision-free path from $\mathbf{q}_{\text{start}}$ to $\mathbf{q}_{\text{goal}}$. Rather than predicting a single deterministic path, \name learns a conditional distribution over joint configuration increments:
\begin{equation} \label{eq - path_dist}
    p_\theta(\delta \mathbf{q}_{t:t+H-1} \mid \mathbf{q}_t, \mathbf{q}_{\text{goal}}, \mathcal{P}),
\end{equation}
where $\mathbf{q}_t$ is the current configuration, $H$ is the planning horizon, and $\theta$ denotes the planner parameters. Trajectories are generated autoregressively by predicting increments:
\begin{equation} \label{eq - autoregressive-planning}
    \delta \mathbf{q}_{t:t+H-1} = \pi_\theta(\mathbf{q}_t, \mathbf{q}_{\text{goal}}, \mathcal{P}).
\end{equation}
We learn this distribution using flow matching.

\subsection{Flow Matching as Motion Policy} \label{subsec: flow_policy}
To model the conditional path distribution, we learn a flow-based policy that is learned under the flow matching framework \cite{lipman2022flow} to model the distribution of planning data $p_0(x^0)$, where $x^0=\delta \mathbf{q}_{t:t+H-1}$. This is achieved by constructing a Continuous Normalizing Flow \cite{chen2018neural}. Specifically, we define the optimal transport probability path:
\begin{equation} \label{eq - optimaltransport}
\begin{aligned}
    x^\tau = (1 - \tau)x^0 + \tau \epsilon,~~~\tau \in [0, 1],
\end{aligned}
\end{equation}
which linearly interpolates between clean data $x^0$ at $\tau = 0$ and Gaussian noise $\epsilon \sim \mathcal{N}(0, I)$ at $\tau = 1$ thereby defining the conditional probability path $p_\tau (x^\tau | x^0)$. The corresponding conditional generating vector field is given by:
\begin{equation} \label{eq - vectorfield}
\begin{aligned}
    u_\tau (x^\tau) = \mathbb{E}_{p(x^0 | x^\tau)} u_\tau (x^\tau | x^0), 
\end{aligned}
\end{equation}
where the conditional vector field $u_\tau (x^\tau | x^0) := \frac{d}{d\tau} x^\tau = \epsilon - x^0$ can be computed directly from samples $x^0$ and $\epsilon$. The model learns an estimator $v_\theta(x^\tau, \tau, \mathcal{O})$ conditioned on planning observation, by regressing it toward the conditional vector field:
\begin{equation} \label{eq - loss}
\begin{aligned}
    \mathcal{L} = \mathbb{E}_{\mathcal{T}(\tau), p_0(x^0), p_\tau (x^\tau | x^0)} \lVert v_\theta (x^\tau, \tau, \mathcal{O}) - u_\tau (x^\tau | x^0)\rVert^2, 
\end{aligned}
\end{equation}
where $\mathcal{T}$ denotes the uniform distribution $\mathcal{U}([0, 1])$ over flow times~\cite{lipman2022flow}. \name learning procedure is demonstrated in Algorithm~\ref{alg:training}.

\begin{algorithm}[thbp]
\caption{\small \name Training}
\label{alg:training}
    \DontPrintSemicolon
    \KwIn{observations $\mathcal{O}$, target increment waypoints $\mathbf{x}^0$}
    \KwIn{vector field estimator $v_\theta$, learning rate $\eta$}
    \While{not converged}{
        \Comment*[l]{Sample waypoints}
        $\epsilon \sim \mathcal{N}(0, I)$ \;
        \Comment*[l]{Sample time steps}
        $t \sim \mathcal{U}[0, 1]$ \;
        \Comment*[l]{Linear interpolation}
        $\mathbf{x}_t = (1-t)\mathbf{x}^0 + t\epsilon$ \;
        \Comment*[l]{Flow estimation}
        $\mathbf{v}_t(\mathbf{x}|\mathcal{O}) = v_\theta (\mathbf{x}_t, t, \mathcal{O})$ \;
        \Comment*[l]{Loss function}
        $\mathcal{L}(\theta) = \lVert v_\theta (\mathbf{x}_t, t, \mathcal{O}) - (\epsilon - x^0) \rVert_2^2$ \;
        \Comment*[l]{Gradient optimization step}
        $\theta = \theta - \eta \nabla_\theta \mathcal{L}(\theta)$\;
    }
    \textbf{Output:} optimized $\theta$ \;
\end{algorithm}%

During inference, samples are obtained by integrating the learned vector field $v_\theta(x^\tau, \tau, \mathcal{O})$ from $\tau = 1$ to $\tau = 0$ to recover $\hat{x}^0 \sim p_0$:
\begin{equation} \label{eq - inference}
\begin{aligned}
    \hat{x}^0 = \epsilon + \int_1^0 v_\theta (\hat{x}^\tau, \tau, \mathcal{O})~d\tau. 
\end{aligned}
\end{equation}

\name is evaluated autoregressively in an open-loop manner. The planning policy is rolled out for a fixed number of steps ($T$), where each predicted joint action chunk is auto-regressively fed back as input for subsequent action joint predictions. Planning success is defined as the end-effector reaching within a predefined threshold of the planning goal pose configuration. Algorithm~\ref{alg:inference} details the inference procedure of \name.

\begin{algorithm}[htbp]
\caption{\small \name Inference}
\label{alg:inference}
\DontPrintSemicolon
    \KwIn{roll-out length $T$, chunk size $H$, flow policy $\mathbf{v}_\theta$}
    \KwIn{number of paths $N$}
    \KwIn{observations $\mathcal{O}^{1:N}$, start configuration $\mathbf{q}_0^{1:N}$}

    \BlankLine
    $\mathcal{Q} \in \mathbb{R}^{N \times (T+1) \times d}~\leftarrow~\emptyset$ \;
    $\mathcal{Q} \gets \{\mathbf{q}_0^{1:N}$\} \;
    $\mathbf{q}_t^{1:N} \leftarrow \mathbf{q}_0^{1:N}$\;

    \BlankLine
    \Comment*[l]{Rolling policy autoregressively}
    \For{$t\leftarrow 0$ \KwTo $T$}{
        \Comment*[l]{Sample noisy waypoints}
        $\epsilon^{1:N} \sim \mathcal{N}(0,\mathbf{I})$ \;
        \Comment*[l]{Flow integration}
        $\delta \mathbf{q}_{t:t+H-1}^{1:N} \gets \epsilon^{1:N} + \int_{1}^{0} \mathbf{v}_{\theta}(\mathbf{x}^{\tau}, \tau \mid \mathcal{O}^{1:N})\, d\tau$ \;
        \vspace{0.3em}
        \Comment*[l]{Rolling over planning horizon}
        \For{$\delta\mathbf{q}^{1:N}$ in $\delta \mathbf{q}_{t:t+H-1}^{1:N}$}{
            \vspace{0.3em}
            $\mathbf{q}_t^{1:N}~\leftarrow~\mathbf{q}_t^{1:N} + \delta\mathbf{q}^{1:N}$ \;
            \vspace{0.3em}
            $\mathcal{Q}~\leftarrow~\{\mathbf{q}_t^{1:N}\}$ \;
        }
    }

    \BlankLine
    \Comment*[l]{Reaching goal}
    $\mathcal{Q}^*~\leftarrow~\emptyset$ \;
    \For{$\mathbf{q}$ in $\mathcal{Q}$}{
        \Comment*[l]{Checking reaching goal condition}
        \If{\textit{ReachedGoal}($\mathbf{q}$)}{
            $\mathcal{Q}^*~\leftarrow~\{\mathbf{q}\}$
        }
    }
    \KwOut{$\mathcal{Q}^*$}
\end{algorithm}

\subsection{Best-of-$N$ Sampling} \label{subsec: best_of_n}
We augment \name with a lightweight inference-time optimization to improve the planning efficiency of the base policy~\cite{dalal2024neural}. Specifically, leveraging the stochastic generative formulation of flow matching~\cite{lipman2022flow}, we sample $N$ candidate paths connecting the initial and goal configurations to generate a set of candidate trajectories $\mathcal{Q}^* = \{q_i\}_{i=1}^{N}$, where each trajectory $q_i = \{\mathbf{q}_t^{(i)}\}_{t=1}^{T}$ represents a sequence of robot configurations over the pre-defined roll-out length $T$. We evaluate each trajectory using a collision checker and select the first collision-free path:
\begin{equation} \label{eq - best-of-n}
\begin{aligned}
    q^* = \arg\min_{q_i}\mathcal{C}(q_i),
\end{aligned}
\end{equation}
where the cost function $\mathcal{C}(q_i)$ is defined as:
\begin{equation} \label{eq - best-of-n-cost}
\begin{aligned}
    \mathcal{C}(q_i) = \sum_{t=1}^T \mathbb{I}\{d(\mathbf{q}_t^i, \mathcal{P}) < \delta_\text{safe}\},
\end{aligned}
\end{equation}
where $d(\mathbf{q}_t^i, \mathcal{P})$ denotes the minimum signed distance between the robot at configuration $\mathbf{q}_t^i$ and the environment representation $\mathcal{P}$, $\mathbb{I}(.)$ is an indicator function, and $\delta_\text{safe}$ is the safety distance threshold. In real-world deployment, $d(\mathbf{q}_t^i, \mathcal{P})$ is computed using robot signed distance function over environment representation. In the simulation, the efficient collision utilities of cuRobo~\cite{sundaralingam2023curobo} are leveraged for batched collision checking of all candidate paths. Algorithm~\ref{alg:best_of_n} details the best-of-$N$ sampling procedure.

\begin{algorithm}[htbp]
\caption{\small Best-of-$N$ Sampling}
\label{alg:best_of_n}
\DontPrintSemicolon
    \KwIn{path set $\mathcal{Q}^*$}
    \KwOut{first collision free path $q^*$}

    \BlankLine
    \For{$q$ in $\mathcal{Q}^*$}{
        $C(q) \gets 0$\;

        \BlankLine
        \For{$\mathbf{q}_t^i$ in $q$}{
            \Comment*[l]{Compute distance}
            $\delta~\leftarrow~d(\mathbf{q}_t^{i}, \mathcal{P})$\;
            \Comment*[l]{Compute cost}
            \If{$\delta < \delta_{\text{safe}}$}{
                $\mathcal{C}(q) \gets \mathcal{C}(q) + 1$\;
            }
        }

        \BlankLine
        \Comment*[l]{Find first collision free path}
        \If{$\mathcal{C}(q) = 0$}{
            $q^* \gets q$ \;
            \Return{$q^*$} 
        }
    }
\end{algorithm}

\subsection{Network Architecture} \label{subsec: network_architecture}

\vspace{0.2cm}
\noindent
\textbf{Embedding robot and workspace planning information.}
The current and goal configurations are embedded via a shared multi-layer perceptron (MLP):
\begin{equation} \label{eq - current}
\begin{aligned}
    \mathbf{h}_t = \text{MLP}(\mathbf{q}_t),~\mathbf{h}_{goal} = \text{MLP}(\mathbf{q}_{goal}),
\end{aligned}
\end{equation}
where $\mathbf{q}_t \in \mathbb{R}^6$, $\mathbf{q}_{goal} \in \mathbb{R}^6$, $\mathbf{h}_t \in \mathbb{R}^d$, and $\mathbf{h}_{goal} \in \mathbb{R}^d$ denote the configuration at time step $t$, the goal configuration, and their corresponding embeddings, respectively, and $d$ is the embedding size. Also, robot and scene point clouds are down-sampled and embedded with set abstraction layers from PointNet++~\cite{qi2017pointnet++}:
\begin{equation} \label{eq - pcd_robot}
\begin{aligned}
    \mathbf{h}_r = \text{PointNet++}(\mathcal{P}_r),~\mathbf{h}_{w} = \text{PointNet++}(\mathcal{P}_w),
\end{aligned}
\end{equation}
where $\mathcal{P}_r \in \mathbb{R}^{N_r \times 3}$ is the point cloud of robotic manipulator, $\mathcal{P}_w \in \mathbb{R}^{N_w \times 3}$ is the point cloud of planning scene, $\mathbf{h}_r \in \mathbb{R}^{K_r \times d}$, and $\mathbf{h}_w \in \mathbb{R}^{K_w \times d}$ are their corresponding embeddings, respectively.

\vspace{0.2cm}
\noindent
\textbf{Transformer encoder.} Each of the planning embeddings, including ($\mathbf{h}_t,~\mathbf{h}_\text{goal},~\mathbf{h}_r,~\mathbf{h}_w$), corresponds to elements of planning observation $\mathbf{o} = \{\mathbf{q}_t, \mathbf{q_\text{goal}, \mathcal{P}_r, \mathcal{P}_w}\}$. These embeddings are augmented with a learnable token embedding: $e_t,~e_\text{goal},~e_r,~e_w \in \mathbb{R}^d$, which is added to each respective feature to encode its semantic role. The resulting representations are then treated as tokens and fed into the transformer encoder input.

\vspace{0.2cm}
\noindent
\textbf{Flow head.} The flow head is instantiated as a multi-layer perceptron (MLP) that operates on learnable action tokens $\hat{\mathbf{x}}^0 \in \mathbb{R}^H$. The output of transformer encoder is also fed into the flow head to enable the model to incorporate planning information. The MLP predicts the continuous-time vector field $\mathbf{v}_t$, which is subsequently used to generate actions through flow-based sampling.

\setlength{\tabcolsep}{2pt}
\begin{table*}[!t]
\begin{center}
\captionof{table}{\small \textbf{Comparison with Bechmark Planners.} Planning performance of \name with 20 inference flow steps and benchmark planners across evaluation planning tasks. ``\textbf{FMP-1}'' denotes Flow Motion Policy without inference-time optimization ($N=1$), while ``\textbf{FMP-100}'' refers to the variant with inference-time optimization using 100 planned paths ($N=100$). ``T'' represents planning time, ``SR'' denotes success rate, ``C'' denotes path length, ``S'' denotes path smoothness, and ``J'' denotes path jerk.`` $\downarrow$'' indicates lower values are preferable, and ``$\uparrow$'' indicates higher values are preferable.}
\label{tab:algo_comparison}
\hspace{-0.2cm}
\renewcommand{\arraystretch}{1.1}
\resizebox{\textwidth}{!}{%
\begin{tabular*}{\linewidth}{@{} l @{\extracolsep{\fill}} c c c c c c c c c c}
\toprule
\phantom{X}&\multicolumn{1}{c}{\scriptsize{\textbf{RRTC}~\cite{kuffner2000rrt}}}&\multicolumn{1}{c}{\scriptsize{\textbf{BIT*}~\cite{gammell2020batch}}}&\multicolumn{1}{c}{\scriptsize{\textbf{cuRobo}~\cite{sundaralingam2023curobo}}}&\multicolumn{1}{c}{\scriptsize{\textbf{cuRobo-Vox}~\cite{sundaralingam2023curobo}}}&\multicolumn{1}{c}{\scriptsize{\textbf{MPNets}~\cite{qureshi2020motion}}}&\multicolumn{1}{c}{\scriptsize{\textbf{SIMPNet}~\cite{soleymanzadeh2025simpnet}}}&\multicolumn{1}{c}{\scriptsize{\textbf{GAIDE}~\cite{soleymanzadeh2026gaide}}}&\multicolumn{1}{c}{\scriptsize{\textbf{PerFACT}~\cite{soleymanzadeh2025perfact}}}&\multicolumn{1}{c}{\scriptsize{\textbf{FMP-1 (ours)}}}&\multicolumn{1}{c}{\scriptsize{\textbf{FMP-100 (ours)}}} \\
\toprule
\rowcolor{gray!20}
\multicolumn{11}{c}{{T $(s)\downarrow$}} \\
\midrule
   \scriptsize{\textbf{TableTop}} & \scriptsize{$2.85{\pm}0.97$} & \scriptsize{$3.19{\pm}2.67$} & \scriptsize{$1.97{\pm}1.98$} & \scriptsize{$1.62{\pm}1.37$} & \scriptsize{$2.66{\pm}1.26$} & \scriptsize{$5.68{\pm}10.20$} & \scriptsize{$3.00{\pm}2.58$} & \scriptsize{$0.22{\pm}0.01$} & \scriptsize{$\mathbf{0.16}{\pm}0.01$} & \scriptsize{$0.58{\pm}0.44$} \\
   \midrule
   \scriptsize{\textbf{Box}} & \scriptsize{$0.73{\pm}0.54$} & \scriptsize{$1.40{\pm}1.47$} & \scriptsize{$0.45{\pm}0.92$} & \scriptsize{$0.46{\pm}0.68$} & \scriptsize{$2.41{\pm}1.01$} & \scriptsize{$2.68{\pm}1.66$} & \scriptsize{$2.17{\pm}2.02$} & \scriptsize{$0.23{\pm}0.05$} & \scriptsize{$\mathbf{0.16}{\pm}0.03$} & \scriptsize{$0.38{\pm}0.07$} \\
   \midrule
   \scriptsize{\textbf{Bins}} & \scriptsize{$1.10{\pm}0.67$} & \scriptsize{$1.84{\pm}1.81$} & \scriptsize{$0.17{\pm}0.20$} & \scriptsize{$\mathbf{0.16}{\pm}0.19$} & \scriptsize{$3.45{\pm}1.62$} & \scriptsize{$3.97{\pm}2.52$} & \scriptsize{$3.72{\pm}3.56$} & \scriptsize{$0.24{\pm}0.04$} & \scriptsize{$\mathbf{0.16}{\pm}0.02$} & \scriptsize{$0.43{\pm}0.27$} \\
   \midrule
   \scriptsize{\textbf{Shelf1}} & \scriptsize{$2.71{\pm}0.73$} & \scriptsize{$4.35{\pm}2.38$} & \scriptsize{$0.89{\pm}1.36$} & \scriptsize{$0.48{\pm}1.13$} & \scriptsize{$2.29{\pm}0.01$} & \scriptsize{$3.28{\pm}0.10$} & \scriptsize{$2.99{\pm}1.58$} & \scriptsize{$0.24{\pm}0.03$} & \scriptsize{$\mathbf{0.16}{\pm}0.03$} & \scriptsize{$0.58{\pm}0.49$} \\
   \midrule
   \scriptsize{\textbf{Shelf2}} & \scriptsize{$3.91{\pm}1.12$} & \scriptsize{$5.95{\pm}2.09$} & \scriptsize{$0.48{\pm}1.22$} & \scriptsize{$0.37{\pm}0.63$} & \scriptsize{$2.57{\pm}0.89$} & \scriptsize{$2.96{\pm}1.43$} & \scriptsize{$5.56{\pm}2.57$} & \scriptsize{$0.22{\pm}0.01$} & \scriptsize{$\mathbf{0.15}{\pm}0.03$} & \scriptsize{$0.67{\pm}0.71$} \\
   \midrule
   \scriptsize{\textbf{Shelf3}} & \scriptsize{$4.07{\pm}1.17$} & \scriptsize{$4.84{\pm}3.24$} & \scriptsize{$0.86{\pm}1.29$} & \scriptsize{$0.86{\pm}1.21$} & \scriptsize{$3.43{\pm}1.19$} & \scriptsize{$6.88{\pm}4.78$} & \scriptsize{$4.34{\pm}3.25$} & \scriptsize{$0.22{\pm}0.01$} & \scriptsize{$\mathbf{0.16}{\pm}0.01$} & \scriptsize{$0.81{\pm}0.88$} \\
   \bottomrule
\end{tabular*}}
\end{center}
\begin{center}
\renewcommand{\arraystretch}{1.1}
\resizebox{\textwidth}{!}{%
\begin{tabular*}{\linewidth}{@{} l @{\extracolsep{\fill}} c c c c c c c c c c}
\toprule
\rowcolor{gray!20}
\multicolumn{11}{c}{SR $(\%)\uparrow$} \\
\midrule
   \scriptsize{\textbf{TableTop}} & \scriptsize{$\mathbf{88.0}$\%} & \scriptsize{$71.0$\%} & \scriptsize{$81.0$\%} & \scriptsize{$76.0$\%} & \scriptsize{$41.0$\%} & \scriptsize{$51.0$\%} & \scriptsize{$52.0$\%} & \scriptsize{$58.0$\%} & \scriptsize{$48.0$\%} & \scriptsize{$84.0$\%} \\
   \midrule
   \scriptsize{\textbf{Box}} & \scriptsize{$\mathbf{71.0}$\%} & \scriptsize{$67.0$\%} & \scriptsize{$63.7$\%} & \scriptsize{$63.3$\%} & \scriptsize{$62.0$\%} & \scriptsize{$67.0$\%} & \scriptsize{$65.0$\%} & \scriptsize{$61.3$\%} & \scriptsize{$60.7$\%} & \scriptsize{$66.0$\%} \\
   \midrule
   \scriptsize{\textbf{Bins}} & \scriptsize{$\mathbf{98.3}$\%} & \scriptsize{$93.8$\%} & \scriptsize{$89.5$\%} & \scriptsize{$89.3$\%} & \scriptsize{$84.5$\%} & \scriptsize{$94.2$\%} & \scriptsize{$96.0$\%} & \scriptsize{$84.5$\%} & \scriptsize{$75.5$\%} & \scriptsize{$96.8$\%} \\
   \midrule
   \scriptsize{\textbf{Shelf1}} & \scriptsize{$\mathbf{98.0}$\%} & \scriptsize{$91.0$\%} & \scriptsize{$37.7$\%} & \scriptsize{$39.7$\%} & \scriptsize{$39.0$\%} & \scriptsize{$44.0$\%} & \scriptsize{$55.0$\%} & \scriptsize{$35.7$\%} & \scriptsize{$45.0$\%} & \scriptsize{$79.4$\%} \\
   \midrule
   \scriptsize{\textbf{Shelf2}} & \scriptsize{$\mathbf{82.6}$\%} & \scriptsize{$75.0$\%} & \scriptsize{$75.0$\%} & \scriptsize{$75.3$\%} & \scriptsize{$34.0$\%} & \scriptsize{$35.0$\%} & \scriptsize{$44.0$\%} & \scriptsize{$34.6$\%} & \scriptsize{$34.4$\%} & \scriptsize{$57.7$\%} \\
   \midrule
   \scriptsize{\textbf{Shelf3}} & \scriptsize{$45.0$\%} & \scriptsize{$38.0$\%} & \scriptsize{$65.7$\%} & \scriptsize{$\mathbf{66.0}$\%} & \scriptsize{$32.0$\%} & \scriptsize{$33.0$\%} & \scriptsize{$38.0$\%} & \scriptsize{$33.4$\%} & \scriptsize{$33.4$\%} & \scriptsize{$53.4$\%} \\
   \bottomrule
\end{tabular*}}
\end{center}

\begin{center}
\renewcommand{\arraystretch}{1.1}
\resizebox{\textwidth}{!}{%
\begin{tabular*}{\linewidth}{@{} l @{\extracolsep{\fill}} c c c c c c c c c c}
\toprule
\rowcolor{gray!20}
\multicolumn{11}{c}{C (rad) $\downarrow$} \\
\midrule
   \scriptsize{\textbf{TableTop}} & \scriptsize{$15.24{\pm}5.58$} & \scriptsize{$9.31{\pm}4.71$} & \scriptsize{$6.06{\pm}2.11$} & \scriptsize{$6.01{\pm}2.22$} & \scriptsize{$5.64{\pm}1.96$} & \scriptsize{$5.67{\pm}2.14$} & \scriptsize{$5.56{\pm}1.91$} & \scriptsize{$\mathbf{5.35}{\pm}1.88$} & \scriptsize{$5.50{\pm}1.98$} & \scriptsize{$6.07{\pm}2.17$} \\
   \midrule
   \scriptsize{\textbf{Box}} & \scriptsize{$12.23{\pm}3.28$} & \scriptsize{$5.19{\pm}2.34$} & \scriptsize{$4.57{\pm}1.47$} & \scriptsize{$4.57{\pm}1.48$} & \scriptsize{$4.84{\pm}1.62$} & \scriptsize{$4.57{\pm}1.47$} & \scriptsize{$4.55{\pm}1.50$} & \scriptsize{$4.87{\pm}1.74$} & \scriptsize{$\mathbf{4.51}{\pm}1.52$} & \scriptsize{$4.64{\pm}1.62$} \\
   \midrule
   \scriptsize{\textbf{Bins}} & \scriptsize{$12.62{\pm}3.42$} & \scriptsize{$6.62{\pm}2.82$} & \scriptsize{$5.26{\pm}1.94$} & \scriptsize{$\mathbf{5.25}{\pm}1.93$} & \scriptsize{$5.58{\pm}1.88$} & \scriptsize{$5.45{\pm}1.84$} & \scriptsize{$5.39{\pm}1.91$} & \scriptsize{$5.39{\pm}1.84$} & \scriptsize{$5.61{\pm}1.92$} & \scriptsize{$5.44{\pm}1.94$} \\
   \midrule
   \scriptsize{\textbf{Shelf1}} & \scriptsize{$20.04{\pm}8.37$} & \scriptsize{$13.48{\pm}4.75$} & \scriptsize{$7.16{\pm}2.20$} & \scriptsize{$7.10{\pm}2.13$} & \scriptsize{$7.69{\pm}2.17$} & \scriptsize{$6.01{\pm}1.89$} & \scriptsize{$6.64{\pm}1.73$} & \scriptsize{$\mathbf{5.27}{\pm}2.28$} & \scriptsize{$6.92{\pm}2.06$} & \scriptsize{$6.91{\pm}1.90$} \\
   \midrule
   \scriptsize{\textbf{Shelf2}} & \scriptsize{$15.68{\pm}6.04$} & \scriptsize{$9.87{\pm}5.24$} & \scriptsize{$5.53{\pm}1.72$} & \scriptsize{$5.57{\pm}1.76$} & \scriptsize{$5.33{\pm}1.57$} & \scriptsize{$5.04{\pm}1.39$} & \scriptsize{$\mathbf{4.87}{\pm}1.51$} & \scriptsize{$5.39{\pm}1.89$} & \scriptsize{$5.06{\pm}1.56$} & \scriptsize{$5.44{\pm}1.77$} \\
   \midrule
   \scriptsize{\textbf{Shelf3}} & \scriptsize{$13.95{\pm}4.78$} & \scriptsize{$7.85{\pm}3.42$} & \scriptsize{$6.14{\pm}1.48$} & \scriptsize{$6.24{\pm}1.72$} & \scriptsize{$6.20{\pm}1.60$} & \scriptsize{$6.04{\pm}1.47$} & \scriptsize{$5.88{\pm}1.40$} & \scriptsize{$5.93{\pm}1.56$} & \scriptsize{$\mathbf{5.84}{\pm}1.38$} & \scriptsize{$6.04{\pm}1.37$} \\
   \bottomrule
\end{tabular*}}
\end{center}

\begin{center}
\renewcommand{\arraystretch}{1.1}
\resizebox{\textwidth}{!}{%
\begin{tabular*}{\linewidth}{@{} l @{\extracolsep{\fill}} c c c c c c c c c c}
\toprule
\rowcolor{gray!20}
\multicolumn{11}{c}{S (rad step$^{-2}$) $\downarrow$} \\
\midrule
  \scriptsize{\textbf{TableTop}} & \scriptsize{$7.02{\pm}1.62$} & \scriptsize{$4.74{\pm}4.46$} & \scriptsize{$\mathbf{0.01}{\pm}0.01$} & \scriptsize{$\mathbf{0.01}{\pm}0.01$} & \scriptsize{$1.83{\pm}2.27$} & \scriptsize{$1.36{\pm}2.19$} & \scriptsize{$1.71{\pm}2.34$} & \scriptsize{$0.03{\pm}0.01$} & \scriptsize{$0.03{\pm}0.01$} & \scriptsize{$0.03{\pm}0.01$} \\
  \midrule
   \scriptsize{\textbf{Box}} & \scriptsize{$7.66{\pm}1.96$} & \scriptsize{$1.20{\pm}2.98$} & \scriptsize{$\mathbf{0.01}{\pm}0.01$} & \scriptsize{$\mathbf{0.01}{\pm}0.01$} & \scriptsize{$1.00{\pm}1.93$} & \scriptsize{$0.62{\pm}1.57$} & \scriptsize{$0.46{\pm}1.41$} & \scriptsize{$0.02{\pm}0.01$} & \scriptsize{$0.03{\pm}0.01$} & \scriptsize{$0.03{\pm}0.01$} \\
   \midrule
   \scriptsize{\textbf{Bins}} & \scriptsize{$7.41{\pm}1.91$} & \scriptsize{$2.15{\pm}3.96$} & \scriptsize{$\mathbf{0.01}{\pm}0.01$} & \scriptsize{$\mathbf{0.01}{\pm}0.01$} & \scriptsize{$1.34{\pm}2.09$} & \scriptsize{$1.04{\pm}1.96$} & \scriptsize{$0.84{\pm}1.97$} & \scriptsize{$0.03{\pm}0.01$} & \scriptsize{$0.03{\pm}0.01$} & \scriptsize{$0.03{\pm}0.01$} \\
   \midrule
   \scriptsize{\textbf{Shelf1}} & \scriptsize{$6.99{\pm}1.28$} & \scriptsize{$7.47{\pm}4.13$} & \scriptsize{$\mathbf{0.01}{\pm}0.01$} & \scriptsize{$\mathbf{0.01}{\pm}0.01$} & \scriptsize{$2.51{\pm}2.88$} & \scriptsize{$0.88{\pm}1.62$} & \scriptsize{$1.18{\pm}2.64$} & \scriptsize{$0.02{\pm}0.01$} & \scriptsize{$0.03{\pm}0.01$} & \scriptsize{$0.03{\pm}0.01$} \\
   \midrule
   \scriptsize{\textbf{Shelf2}} & \scriptsize{$7.40{\pm}1.59$} & \scriptsize{$4.63{\pm}3.95$} & \scriptsize{$\mathbf{0.01}{\pm}0.01$} & \scriptsize{$\mathbf{0.01}{\pm}0.01$} & \scriptsize{$1.44{\pm}2.31$} & \scriptsize{$0.94{\pm}1.68$} & \scriptsize{$0.83{\pm}1.59$} & \scriptsize{$0.03{\pm}0.01$} & \scriptsize{$0.03{\pm}0.01$} & \scriptsize{$0.03{\pm}0.01$} \\
   \midrule
   \scriptsize{\textbf{Shelf3}} & \scriptsize{$6.97{\pm}1.49$} & \scriptsize{$2.79{\pm}3.96$} & \scriptsize{$\mathbf{0.01}{\pm}0.01$} & \scriptsize{$\mathbf{0.01}{\pm}0.01$} & \scriptsize{$1.15{\pm}2.05$} & \scriptsize{$0.63{\pm}1.41$} & \scriptsize{$0.58{\pm}1.62$} & \scriptsize{$0.03{\pm}0.01$} & \scriptsize{$0.03{\pm}0.01$} & \scriptsize{$0.03{\pm}0.01$} \\
   \bottomrule
\end{tabular*}}
\end{center}

\begin{center}
\renewcommand{\arraystretch}{1.1}
\resizebox{\textwidth}{!}{%
\begin{tabular*}{\linewidth}{@{} l @{\extracolsep{\fill}} c c c c c c c c c c}
\toprule
\rowcolor{gray!20}
\multicolumn{11}{c}{J (rad step$^{-3}$) $\downarrow$} \\
\midrule
   \scriptsize{\textbf{TableTop}} & \scriptsize{$6.03{\pm}4.95$} & \scriptsize{$3.42{\pm}6.52$} & \scriptsize{$\mathbf{0.01}{\pm}0.01$} & \scriptsize{$\mathbf{0.01}{\pm}0.01$} & \scriptsize{$0.41{\pm}1.67$} & \scriptsize{$0.15{\pm}0.70$} & \scriptsize{$0.96{\pm}2.09$} & \scriptsize{$0.02{\pm}0.01$} & \scriptsize{$0.02{\pm}0.01$} & \scriptsize{$0.02{\pm}0.01$} \\
   \midrule
   \scriptsize{\textbf{Box}} & \scriptsize{$3.05{\pm}4.02$} & \scriptsize{$0.07{\pm}0.96$} & \scriptsize{$\mathbf{0.01}{\pm}0.01$} & \scriptsize{$\mathbf{0.01}{\pm}0.01$} & \scriptsize{$0.15{\pm}1.06$} & \scriptsize{$0.16{\pm}1.04$} & \scriptsize{$0.29{\pm}1.54$} & \scriptsize{$0.02{\pm}0.01$} & \scriptsize{$0.02{\pm}0.01$} & \scriptsize{$0.02{\pm}0.01$} \\
   \midrule
   \scriptsize{\textbf{Bins}} & \scriptsize{$3.95{\pm}4.77$} & \scriptsize{$0.12{\pm}1.17$} & \scriptsize{$\mathbf{0.01}{\pm}0.01$} & \scriptsize{$\mathbf{0.01}{\pm}0.01$} & \scriptsize{$0.31{\pm}1.61$} & \scriptsize{$0.39{\pm}1.53$} & \scriptsize{$0.50{\pm}2.01$} & \scriptsize{$0.02{\pm}0.01$} & \scriptsize{$0.02{\pm}0.01$} & \scriptsize{$0.02{\pm}0.01$} \\
   \midrule
   \scriptsize{\textbf{Shelf1}} & \scriptsize{$8.89{\pm}4.56$} & \scriptsize{$6.74{\pm}7.76$} & \scriptsize{$0.01{\pm}0.01$} & \scriptsize{$0.01{\pm}0.01$} & \scriptsize{$1.58{\pm}2.58$} & \scriptsize{$\mathbf{0.00}{\pm}0.00$} & \scriptsize{$0.80{\pm}2.46$} & \scriptsize{$0.02{\pm}0.01$} & \scriptsize{$0.03{\pm}0.01$} & \scriptsize{$0.03{\pm}0.01$} \\
   \midrule
   \scriptsize{\textbf{Shelf2}} & \scriptsize{$6.60{\pm}5.56$} & \scriptsize{$4.78{\pm}6.32$} & \scriptsize{$\mathbf{0.01}{\pm}0.01$} & \scriptsize{$\mathbf{0.01}{\pm}0.01$} & \scriptsize{$0.18{\pm}1.04$} & \scriptsize{$0.58{\pm}1.68$} & \scriptsize{$0.58{\pm}1.38$} & \scriptsize{$0.02{\pm}0.01$} & \scriptsize{$0.02{\pm}0.01$} & \scriptsize{$0.03{\pm}0.01$} \\
   \midrule
   \scriptsize{\textbf{Shelf3}} & \scriptsize{$5.15{\pm}5.10$} & \scriptsize{$1.29{\pm}3.62$} & \scriptsize{$\mathbf{0.01}{\pm}0.01$} & \scriptsize{$\mathbf{0.01}{\pm}0.01$} & \scriptsize{$0.58{\pm}2.05$} & \scriptsize{$0.56{\pm}1.67$} & \scriptsize{$0.44{\pm}1.76$} & \scriptsize{$0.02{\pm}0.01$} & \scriptsize{$0.03{\pm}0.01$} & \scriptsize{$0.03{\pm}0.01$} \\
  \bottomrule
\end{tabular*}}
\end{center}
\end{table*}

\section{Evaluation} \label{sec: results}
We evaluate \name against representative sampling-based and neural motion planning methods. All models are implemented in PyTorch~\cite{paszke2017automatic}, and experiments are run on an NVIDIA RTX 4080 GPU.

\subsection{\name Training}
We use an LLM-based workspace generator to create motion planning environments and use cuRobo~\cite{sundaralingam2023curobo} to generate expert trajectories. The training dataset contains about 3.5 million trajectories from about 500 generated workspaces~\cite{soleymanzadeh2025perfact}. We use the same dataset and training setup for Flow Motion Policy, the neural baselines, and all ablation models to ensure a fair comparison.

\subsection{Benchmarks and Planning Metrics}
\noindent
\textbf{Benchmarks.} We compare the performance of \name with a set of representative sampling-based and neural motion planning approaches. The benchmarks considered in this study are summarized as follows:

We compare \name against representative sampling-based, optimization-based, and neural motion planning baselines. The sampling-based baselines include RRT-Connect (RRTC)~\cite{kuffner2000rrt} and BIT*~\cite{gammell2020batch}, both evaluated using their OMPL implementations~\cite{sucan2012open} with timeouts matched to the neural sampler baselines. We also compare against cuRobo~\cite{sundaralingam2023curobo}, a GPU-accelerated motion generation framework, using both mesh-based (cuRobo) and voxel-based (cuRobo-Vox) collision representations. Neural baselines include MPNets~\cite{qureshi2020motion}, SIMPNet~\cite{soleymanzadeh2025simpnet}, and GAIDE~\cite{soleymanzadeh2026gaide}, which use learned samplers to guide motion planning, as well as PerFACT~\cite{soleymanzadeh2025perfact}, an end-to-end neural planner based on action-chunking transformer fusion.

\vspace{0.2cm}
\noindent
\textbf{Metrics.} We assess the performance of \name using five standard evaluation metrics: \textit{planning time}, and \textit{success rate}, \textit{planning cost}, \textit{smoothness}, and \textit{jerk}. \textit{Planning time} ``T'' measures the average time required to generate a path for each evaluation task, while \textit{Success rate} ``SR'' demonstrates the proportion of solved planning problems. \textit{Planning cost} ``C'' measures the total path length in configuration space.  \textit{Smoothness} ``S'' measures the average change in velocity along the generated path.  \textit{Jerk} ``J'' measures the average change in acceleration along the path.

\begin{figure*}[t]
    \includegraphics[width=0.8\linewidth]{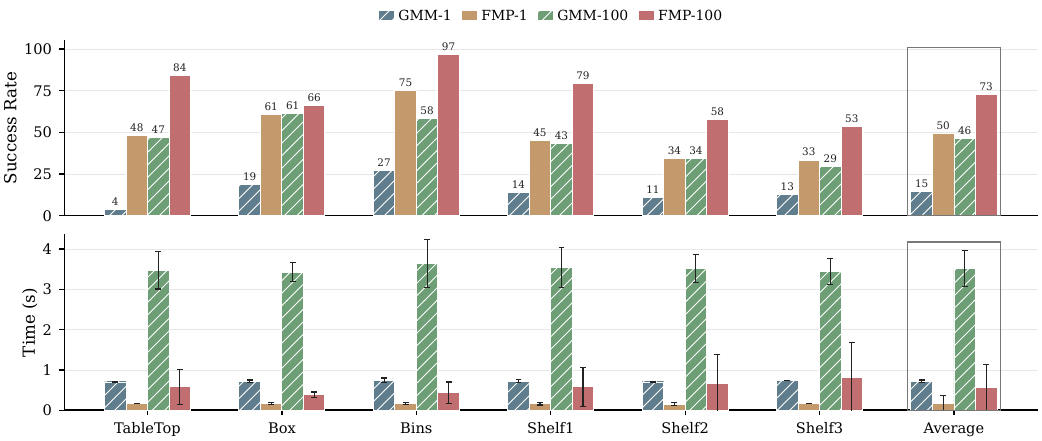}
    \centering
    \captionof{figure}{\small \textbf{Best-of-$\mathbf{N}$ Sampling.} Planning success rate and planning time of \name and Neural MP~\cite{dalal2024neural} adaptation across the held-out planning tasks. ``GMM-1'' and ``GMM-100'' are Neural MP~\cite{dalal2024neural} adaptation with Gaussian Mixture Model (GMM) without inference-time optimization ($N=1$) and inference-time optimization ($N=100$), respectively. ``FMP-1'' and ``FMP-100'' are \name without inference-time optimization and inference-time optimization ($N=100$), respectively.}
    \label{fig:n_sampling}
\end{figure*}

\setlength{\tabcolsep}{2pt}
\begin{table*}[!t]
\begin{center}
\captionof{table}{\small \textbf{Policy Head Architecture Ablation.} Planning success rate and planning time of \name with 20 inference flow steps and Diffusion Motion Policy with 100 denoising steps during inference with various head architectures. ``DMP-1'' and ``DMP-100'' are Diffusion Motion Policy without inference-time optimization and inference-time optimization ($N=100$), respectively. ``FMP-1'' and ``FMP-100'' are \name without inference-time optimization and inference-time optimization ($N=100$), respectively. ``DiT'' denotes DiT-Block Policy~\cite{dasari2025ingredients}. ``T'' represents planning time, ``SR'' denotes success rate, ``C'' denotes path length, ``S'' denotes path smoothness, and ``J'' denotes path jerk.`` $\downarrow$'' indicates lower values are preferable, and ``$\uparrow$'' indicates higher values are preferable.}

\label{tab:head_ablation}

\hspace{-0.2cm}
\renewcommand{\arraystretch}{1.1}
\resizebox{\textwidth}{!}{%
\begin{tabular*}{\textwidth}{@{}l
@{\extracolsep{\fill}}
c c c c c c c@{\extracolsep{\fill}} c
c c c c c c}
\toprule
\rowcolor{gray!20}
\multicolumn{14}{c}{\textbf{\name ($N=1$)}} \\
\midrule
\phantom{Var.} &  
\multicolumn{2}{c}{\scriptsize{\textbf{TableTop}}}&\multicolumn{2}{c}{\scriptsize{\textbf{Box}}}&\multicolumn{2}{c}{\scriptsize{\textbf{Bins}}} && \multicolumn{6}{c}{\scriptsize{\textbf{Shelf}}}\\
\cmidrule{2-3}
\cmidrule{4-5}
\cmidrule{6-7}
\cmidrule{9-14}
\phantom{Var.}&\multicolumn{6}{c}{\phantom{Var.}}&&\multicolumn{2}{c}{\scriptsize{\textbf{Task I}}}&\multicolumn{2}{c}{\scriptsize{\textbf{Task II}}}&\multicolumn{2}{c}{\scriptsize{\textbf{Task III}}} \\
\cmidrule{9-14}
& \scriptsize{T $(s)\downarrow$} & \scriptsize{SR $(\%)\uparrow$} & \scriptsize{T $(s)\downarrow$} & \scriptsize{SR $(\%)\uparrow$} & \scriptsize{T $(s)\downarrow$} & \scriptsize{SR $(\%)\uparrow$} & {} & \scriptsize{T $(s)\downarrow$} & \scriptsize{SR $(\%)\uparrow$}& \scriptsize{T $(s)\downarrow$} & \scriptsize{SR $(\%)\uparrow$}& \scriptsize{T $(s)\downarrow$} & \scriptsize{SR $(\%)\uparrow$} \\
\toprule
\scriptsize{\textbf{MLP}}&\scriptsize{$\mathbf{0.16 {\pm} 0.01}$}&\scriptsize{$\mathbf{48.0}$\%}&\scriptsize{$\mathbf{0.16 {\pm} 0.02}$}&\scriptsize{$60.7$\%}&\scriptsize{$\mathbf{0.16 {\pm} 0.03}$}&\scriptsize{$75.5$\%}&&\scriptsize{$\mathbf{0.16 {\pm} 0.03}$}&\scriptsize{$45.0$\%}&\scriptsize{$\mathbf{0.15 {\pm} 0.03}$}&\scriptsize{$34.4$\%}&\scriptsize{$\mathbf{0.16 {\pm} 0.01}$}&\scriptsize{$33.4$\%} \\
\midrule
\scriptsize{\textbf{U-Net} \cite{chi2025diffusion}}&\scriptsize{$0.75 {\pm} 0.01$}&\scriptsize{$\mathbf{48.0}$\%}&\scriptsize{$0.75 {\pm} 0.03$}&\scriptsize{$59.4$\%}&\scriptsize{$0.77 {\pm} 0.03$}&\scriptsize{$72.3$\%}&&\scriptsize{$0.77 {\pm} 0.02$}&\scriptsize{$44.0$\%}&\scriptsize{$0.76 {\pm} 0.03$}&\scriptsize{$31.4$\%}&\scriptsize{$0.77 {\pm} 0.01$}&\scriptsize{$32.4$\%} \\
\midrule
\scriptsize{\textbf{Transformer} \cite{chi2025diffusion}}&\scriptsize{$0.55 {\pm} 0.01$}&\scriptsize{$43.0$\%}&\scriptsize{$0.57 {\pm} 0.01$}&\scriptsize{$60.0$\%}&\scriptsize{$0.54 {\pm} 0.03$}&\scriptsize{$77.3$\%}&&\scriptsize{$0.55 {\pm} 0.01$}&\scriptsize{$53.0$\%}&\scriptsize{$0.56 {\pm} 0.03$}&\scriptsize{$\mathbf{36.0}$\%}&\scriptsize{$0.55 {\pm} 0.02$}&\scriptsize{$\mathbf{35.0}$\%} \\
\midrule
\scriptsize{\textbf{DiT} \cite{dasari2025ingredients}}&\scriptsize{$0.57 {\pm} 0.02$}&\scriptsize{$47.0$\%}&\scriptsize{$0.55 {\pm} 0.03$}&\scriptsize{$\mathbf{62.0}$\%}&\scriptsize{$0.55 {\pm} 0.03$}&\scriptsize{$\mathbf{80.0}$\%}&&\scriptsize{$0.54 {\pm} 0.03$}&\scriptsize{$\mathbf{56.4}$\%}&\scriptsize{$0.56 {\pm} 0.03$}&\scriptsize{$34.7$\%}&\scriptsize{$0.56 {\pm} 0.01$}&\scriptsize{$34.0$\%} \\
\bottomrule
\end{tabular*}}
\end{center}
\begin{center}
\renewcommand{\arraystretch}{1.1}
\resizebox{\textwidth}{!}{%
\begin{tabular*}{\linewidth}{@{}l
@{\extracolsep{\fill}}
c c c c c c c@{\extracolsep{\fill}} c
c c c c c c}
\toprule
\rowcolor{gray!20}
\multicolumn{14}{c}{\textbf{Diffusion Motion Policy ($N=1$)}} \\
\midrule
\phantom{Var.} &  
\multicolumn{2}{c}{\scriptsize{\textbf{TableTop}}}&\multicolumn{2}{c}{\scriptsize{\textbf{Box}}}&\multicolumn{2}{c}{\scriptsize{\textbf{Bins}}} && \multicolumn{6}{c}{\scriptsize{\textbf{Shelf}}}\\
\cmidrule{2-3}
\cmidrule{4-5}
\cmidrule{6-7}
\cmidrule{9-14}
\phantom{Var.}&\multicolumn{6}{c}{\phantom{Var.}}&&\multicolumn{2}{c}{\scriptsize{\textbf{Task I}}}&\multicolumn{2}{c}{\scriptsize{\textbf{Task II}}}&\multicolumn{2}{c}{\scriptsize{\textbf{Task III}}} \\
\cmidrule{9-14}
& \scriptsize{T $(s)\downarrow$} & \scriptsize{SR $(\%)\uparrow$} & \scriptsize{T $(s)\downarrow$} & \scriptsize{SR $(\%)\uparrow$} & \scriptsize{T $(s)\downarrow$} & \scriptsize{SR $(\%)\uparrow$} & {} & \scriptsize{T $(s)\downarrow$} & \scriptsize{SR $(\%)\uparrow$}& \scriptsize{T $(s)\downarrow$} & \scriptsize{SR $(\%)\uparrow$}& \scriptsize{T $(s)\downarrow$} & \scriptsize{SR $(\%)\uparrow$} \\
\toprule
\scriptsize{\textbf{MLP}}&\scriptsize{$\mathbf{0.37 {\pm} 0.02}$}&\scriptsize{$34.0$\%}&\scriptsize{$\mathbf{0.38 {\pm} 0.01}$}&\scriptsize{$55.7$\%}&\scriptsize{$\mathbf{0.38 {\pm} 0.02}$}&\scriptsize{$57.3$\%}&&\scriptsize{$\mathbf{0.38 {\pm} 0.01}$}&\scriptsize{$35.4$\%}&\scriptsize{$\mathbf{0.37 {\pm} 0.01}$}&\scriptsize{$26.0$\%}&\scriptsize{$\mathbf{0.37 {\pm} 0.02}$}&\scriptsize{$31.0$\%} \\
\midrule
\scriptsize{\textbf{U-Net} \cite{chi2025diffusion}}&\scriptsize{$2.08 {\pm} 0.05$}&\scriptsize{$\mathbf{49.0}$\%}&\scriptsize{$2.75 {\pm} 0.03$}&\scriptsize{$55.7$\%}&\scriptsize{$2.78 {\pm} 0.04$}&\scriptsize{$62.0$\%}&&\scriptsize{$2.63 {\pm} 0.03$}&\scriptsize{$34.4$\%}&\scriptsize{$2.65 {\pm} 0.04$}&\scriptsize{$26.7$\%}&\scriptsize{$2.75 {\pm} 0.05$}&\scriptsize{$28.7$\%} \\
\midrule
\scriptsize{\textbf{Transformer} \cite{chi2025diffusion}}&\scriptsize{$2.31 {\pm} 0.02$}&\scriptsize{$43.0$\%}&\scriptsize{$2.28 {\pm} 0.04$}&\scriptsize{$\mathbf{59.0}$\%}&\scriptsize{$2.31 {\pm} 0.10$}&\scriptsize{$70.0$\%}&&\scriptsize{$2.21 {\pm} 0.03$}&\scriptsize{$\mathbf{38.0}$\%}&\scriptsize{$2.17 {\pm} 0.05$}&\scriptsize{$30.7$\%}&\scriptsize{$2.29 {\pm} 0.01$}&\scriptsize{$\mathbf{32.7}$\%} \\
\midrule
\scriptsize{\textbf{DiT} \cite{dasari2025ingredients}}&\scriptsize{$2.28 {\pm} 0.07$}&\scriptsize{$45.0$\%}&\scriptsize{$2.22 {\pm} 0.05$}&\scriptsize{$58.4$\%}&\scriptsize{$2.25 {\pm} 0.05$}&\scriptsize{$\mathbf{70.3}$\%}&&\scriptsize{$2.20 {\pm} 0.04$}&\scriptsize{$35.7$\%}&\scriptsize{$2.26 {\pm} 0.05$}&\scriptsize{$\mathbf{32.7}$\%}&\scriptsize{$2.15 {\pm} 0.02$}&\scriptsize{$30.4$\%} \\
\bottomrule
\end{tabular*}} 
\end{center}
\begin{center}
\renewcommand{\arraystretch}{1.1}
\resizebox{\textwidth}{!}{%
\begin{tabular*}{\linewidth}{@{}l
@{\extracolsep{\fill}}
c c c c c c c@{\extracolsep{\fill}} c
c c c c c c}
\toprule
\rowcolor{gray!20}
\multicolumn{14}{c}{\textbf{\name ($N=100$)}} \\
\midrule
\phantom{Var.} &  
\multicolumn{2}{c}{\scriptsize{\textbf{TableTop}}}&\multicolumn{2}{c}{\scriptsize{\textbf{Box}}}&\multicolumn{2}{c}{\scriptsize{\textbf{Bins}}} && \multicolumn{6}{c}{\scriptsize{\textbf{Shelf}}}\\
\cmidrule{2-3}
\cmidrule{4-5}
\cmidrule{6-7}
\cmidrule{9-14}
\phantom{Var.}&\multicolumn{6}{c}{\phantom{Var.}}&&\multicolumn{2}{c}{\scriptsize{\textbf{Task I}}}&\multicolumn{2}{c}{\scriptsize{\textbf{Task II}}}&\multicolumn{2}{c}{\scriptsize{\textbf{Task III}}} \\
\cmidrule{9-14}
& \scriptsize{T $(s)\downarrow$} & \scriptsize{SR $(\%)\uparrow$} & \scriptsize{T $(s)\downarrow$} & \scriptsize{SR $(\%)\uparrow$} & \scriptsize{T $(s)\downarrow$} & \scriptsize{SR $(\%)\uparrow$} & {} & \scriptsize{T $(s)\downarrow$} & \scriptsize{SR $(\%)\uparrow$}& \scriptsize{T $(s)\downarrow$} & \scriptsize{SR $(\%)\uparrow$}& \scriptsize{T $(s)\downarrow$} & \scriptsize{SR $(\%)\uparrow$} \\
\toprule
\scriptsize{\textbf{MLP}}&\scriptsize{$\mathbf{0.58 {\pm} 0.44}$}&\scriptsize{$84.0$\%}&\scriptsize{$\mathbf{0.38 {\pm} 0.07}$}&\scriptsize{$66.0$\%}&\scriptsize{$\mathbf{0.43 {\pm} 0.27}$}&\scriptsize{$\mathbf{96.8}$\%}&&\scriptsize{$\mathbf{0.58 {\pm} 0.49}$}&\scriptsize{$79.4$\%}&\scriptsize{$\mathbf{0.67 {\pm} 0.71}$}&\scriptsize{$57.7$\%}&\scriptsize{$\mathbf{0.81 {\pm} 0.88}$}&\scriptsize{$53.4$\%} \\
\midrule
\scriptsize{\textbf{U-Net} \cite{chi2025diffusion}}&\scriptsize{$1.23 {\pm} 0.43$}&\scriptsize{$\mathbf{85.0}$\%}&\scriptsize{$1.14 {\pm} 0.47$}&\scriptsize{$68.0$\%}&\scriptsize{$1.07 {\pm} 0.23$}&\scriptsize{$96.3$\%}&&\scriptsize{$1.25 {\pm} 0.61$}&\scriptsize{$85.7$\%}&\scriptsize{$1.38 {\pm} 0.79$}&\scriptsize{$53.7$\%}&\scriptsize{$1.32 {\pm} 0.05$}&\scriptsize{$\mathbf{53.7}$\%} \\
\midrule
\scriptsize{\textbf{Transformer} \cite{chi2025diffusion}}&\scriptsize{$1.23 {\pm} 0.75$}&\scriptsize{$80.0$\%}&\scriptsize{$1.08 {\pm} 0.70$}&\scriptsize{$67.4$\%}&\scriptsize{$0.91 {\pm} 0.26$}&\scriptsize{$95.5$\%}&&\scriptsize{$1.08 {\pm} 0.60$}&\scriptsize{$83.7$\%}&\scriptsize{$1.17 {\pm} 0.70$}&\scriptsize{$56.7$\%}&\scriptsize{$1.28 {\pm} 0.72$}&\scriptsize{$46.4$\%} \\
\midrule
\scriptsize{\textbf{DiT} \cite{dasari2025ingredients}}&\scriptsize{$1.06 {\pm} 0.59$}&\scriptsize{$83.0$\%}&\scriptsize{$0.96 {\pm} 0.62$}&\scriptsize{$\mathbf{68.4}$\%}&\scriptsize{$0.86 {\pm} 0.25$}&\scriptsize{$96.5$\%}&&\scriptsize{$1.0 {\pm} 0.57$}&\scriptsize{$\mathbf{86.4}$\%}&\scriptsize{$1.22 {\pm} 0.90$}&\scriptsize{$\mathbf{58.0}$\%}&\scriptsize{$1.32 {\pm} 0.92$}&\scriptsize{$52.0$\%} \\
\bottomrule
\end{tabular*}} 
\end{center}
\begin{center}
\renewcommand{\arraystretch}{1.1}
\resizebox{\textwidth}{!}{%
\begin{tabular*}{\linewidth}{@{}l
@{\extracolsep{\fill}}
c c c c c c c@{\extracolsep{\fill}} c
c c c c c c}
\toprule
\rowcolor{gray!20}
\multicolumn{14}{c}{\textbf{Diffusion Motion Policy ($N=100$)}} \\
\midrule
\phantom{Var.} &  
\multicolumn{2}{c}{\scriptsize{\textbf{TableTop}}}&\multicolumn{2}{c}{\scriptsize{\textbf{Box}}}&\multicolumn{2}{c}{\scriptsize{\textbf{Bins}}} && \multicolumn{6}{c}{\scriptsize{\textbf{Shelf}}}\\
\cmidrule{2-3}
\cmidrule{4-5}
\cmidrule{6-7}
\cmidrule{9-14}
\phantom{Var.}&\multicolumn{6}{c}{\phantom{Var.}}&&\multicolumn{2}{c}{\scriptsize{\textbf{Task I}}}&\multicolumn{2}{c}{\scriptsize{\textbf{Task II}}}&\multicolumn{2}{c}{\scriptsize{\textbf{Task III}}} \\
\cmidrule{9-14}
& \scriptsize{T $(s)\downarrow$} & \scriptsize{SR $(\%)\uparrow$} & \scriptsize{T $(s)\downarrow$} & \scriptsize{SR $(\%)\uparrow$} & \scriptsize{T $(s)\downarrow$} & \scriptsize{SR $(\%)\uparrow$} & {} & \scriptsize{T $(s)\downarrow$} & \scriptsize{SR $(\%)\uparrow$}& \scriptsize{T $(s)\downarrow$} & \scriptsize{SR $(\%)\uparrow$}& \scriptsize{T $(s)\downarrow$} & \scriptsize{SR $(\%)\uparrow$} \\
\toprule
\scriptsize{\textbf{MLP}}&\scriptsize{$\mathbf{0.74 {\pm} 0.36}$}&\scriptsize{$71.0$\%}&\scriptsize{$\mathbf{0.63 {\pm} 0.34}$}&\scriptsize{$65.7$\%}&\scriptsize{$\mathbf{0.68 {\pm} 0.50}$}&\scriptsize{$90.5$\%}&&\scriptsize{$\mathbf{0.69 {\pm} 0.40}$}&\scriptsize{$69.4$\%}&\scriptsize{$\mathbf{0.88 {\pm} 0.77}$}&\scriptsize{$50.0$\%}&\scriptsize{$\mathbf{0.95 {\pm} 0.79}$}&\scriptsize{$53.4$\%} \\
\midrule
\scriptsize{\textbf{U-Net} \cite{chi2025diffusion}}&\scriptsize{$3.19 {\pm} 0.58$}&\scriptsize{$\mathbf{81.0}$\%}&\scriptsize{$3.09 {\pm} 0.47$}&\scriptsize{$67.7$\%}&\scriptsize{$3.02 {\pm} 0.37$}&\scriptsize{$92.8$\%}&&\scriptsize{$3.16 {\pm} 0.70$}&\scriptsize{$68.4$\%}&\scriptsize{$3.20 {\pm} 0.71$}&\scriptsize{$52.0$\%}&\scriptsize{$3.18 {\pm} 0.55$}&\scriptsize{$48.7$\%} \\
\midrule
\scriptsize{\textbf{Transformer} \cite{chi2025diffusion}}&\scriptsize{$3.02 {\pm} 0.55$}&\scriptsize{$78.0$\%}&\scriptsize{$2.89 {\pm} 0.53$}&\scriptsize{$66.0$\%}&\scriptsize{$2.89 {\pm} 0.37$}&\scriptsize{$90.5$\%}&&\scriptsize{$3.11 {\pm} 0.71$}&\scriptsize{$65.4$\%}&\scriptsize{$3.00 {\pm} 0.44$}&\scriptsize{$46.0$\%}&\scriptsize{$3.04 {\pm} 0.46$}&\scriptsize{$48.0$\%} \\
\midrule
\scriptsize{\textbf{DiT} \cite{dasari2025ingredients}}&\scriptsize{$2.87 {\pm} 0.64$}&\scriptsize{$77.0$\%}&\scriptsize{$2.63 {\pm} 0.32$}&\scriptsize{$\mathbf{68.0}$\%}&\scriptsize{$2.56 {\pm} 0.26$}&\scriptsize{$\mathbf{93.3}$\%}&&\scriptsize{$2.83 {\pm} 0.72$}&\scriptsize{$\mathbf{71.7}$\%}&\scriptsize{$2.75 {\pm} 0.58$}&\scriptsize{$\mathbf{60.0}$\%}&\scriptsize{$2.94 {\pm} 0.67$}&\scriptsize{$\mathbf{53.7}$\%} \\
\bottomrule
\end{tabular*}} 
\end{center}
\end{table*}

\subsection{Results}
\begin{figure}
    \includegraphics[width=0.9\linewidth]{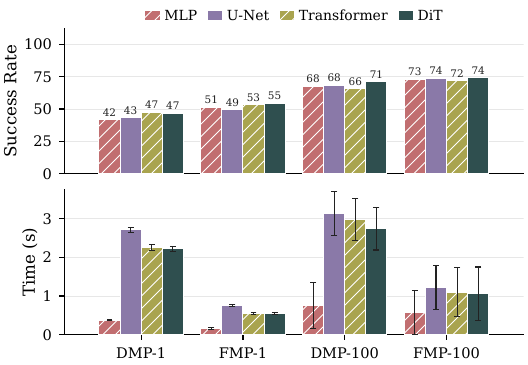}
    \centering
    \captionof{figure}{\small \textbf{Policy Head Architecture Ablation.} Average planning success rate and planning time of \name and Diffusion Motion Policy with various head architectures across all held-out evaluation tasks. ``DMP-1'' and ``DMP-100'' are Diffusion Motion Policy without inference-time optimization ($N=1$) and inference-time optimization ($N=100$), respectively. ``FMP-1'' and ``FMP-100'' are \name without inference-time optimization ($N=1$) and inference-time optimization ($N=100$), respectively. ``DiT'' denotes DiT-Block Policy~\cite{dasari2025ingredients}.}
    \label{fig:ablation}
\end{figure}

We evaluated \name and benchmark planners across held-out planning tasks adopted from recent literature~\cite{dalal2024neural, soleymanzadeh2025perfact}. These evaluation tasks are selected to reflect realistic planning scenarios and be distinct from the generated training workspaces.

\vspace{0.2cm}
\noindent
\textbf{Comparison with benchmarks.} \name with best-of-$N$ sampling achieves success rates comparable to classical sampling-based planners while using much less planning time. It also maintains competitive path quality in terms of cost, smoothness, and jerk as shown in Table~\ref{tab:algo_comparison}. This is because sampling-based planners perform many collision checks during planning, whereas \name efficiently samples candidate trajectories and validates them afterward. Without best-of-$N$ sampling, the base policy is faster but less reliable because a single trajectory may violate planning constraints.

Also, as shown in Table~\ref{tab:algo_comparison}, \name remains competitive with cuRobo and cuRobo-Vox across most evaluation tasks' success rate, planning cost, smoothness, and jerk, with lower planning time. This can be attributed to the fact that cuRobo performs collision checking and trajectory optimization independently for each planning query.

As shown in Table~\ref{tab:algo_comparison}, \name without best-of-$N$ sampling is comparable to neural-informed samplers in success rate, planning cost, smoothness, and jerk, while requiring lower planning time. This is because neural-informed samplers still use collision checking during tree expansion, whereas \name directly predicts trajectories. With best-of-$N$ sampling, \name further improves success rate and path quality while keeping the runtime advantage.

Furthermore, \name without best-of-$N$ sampling performs comparably to PerFACT in success rate, planning cost, smoothness, and jerk, while requiring lower planning time as shown in Table~\ref{tab:algo_comparison}. This is mainly due to the smaller network size of as listed in Table~\ref{tab:ablate-head}. With inference-time optimization, \name further improves success rate and path quality across cost, smoothness, and jerk through stochastic trajectory generation and best-of-$N$ sampling, with only a slight increase in planning time.

\begin{table}[htbp]
    \centering
    \resizebox{0.45\textwidth}{!}{%
    \begin{tabular}{lccccc}
        \toprule
         \textbf{Policy Head}&MLP&U-Net~\cite{chi2025diffusion}&Transformer~\cite{chi2025diffusion}&DiT~\cite{dasari2025ingredients}&Neural MP~\cite{dalal2024neural}  \\
         \midrule
         \textbf{Network Size $\downarrow$}&\textbf{1.40M}&2.29M&2.97M&3.20M&20.00M \\
         \bottomrule
    \end{tabular}}
    \captionof{table}{\small \textbf{Policy Network Size.} Policy parameter size of \name and ablative structures. ``DiT'' denotes DiT-Block Policy~\cite{dasari2025ingredients}, and ``$\downarrow$'' indicates lower is better.}
    \label{tab:ablate-head}
\end{table}

\vspace{0.2cm}
\noindent
\textbf{Best-of-$\mathbf{N}$ sampling.} To evaluate the effectiveness of best-of-$N$ sampling during inference, we adopt and implement Neural MP~\cite{dalal2024neural}. Neural MP utilizes a Gaussian Mixture Model (GMM) for inference-time best-of-$N$ sampling. We adopt this framework based on the official implementation to be compatible with our UR5e robotic manipulator.

As shown in Figure~\ref{fig:n_sampling}, \name consistently outperforms the adapted Neural MP baseline across all held-out tasks. This is mainly due to its flow matching formulation, which can model complex planning distributions more flexibly than the GMM-based Neural MP policy. As a result, \name generates more diverse collision-free candidates for best-of-$N$ sampling. In addition, \name is much lighter, with only \underline{1.40M} parameters compared to Neural MP’s \underline{20.00M}, leading to faster inference and lower computational cost.

\begin{figure}
    \includegraphics[width=0.8\linewidth]{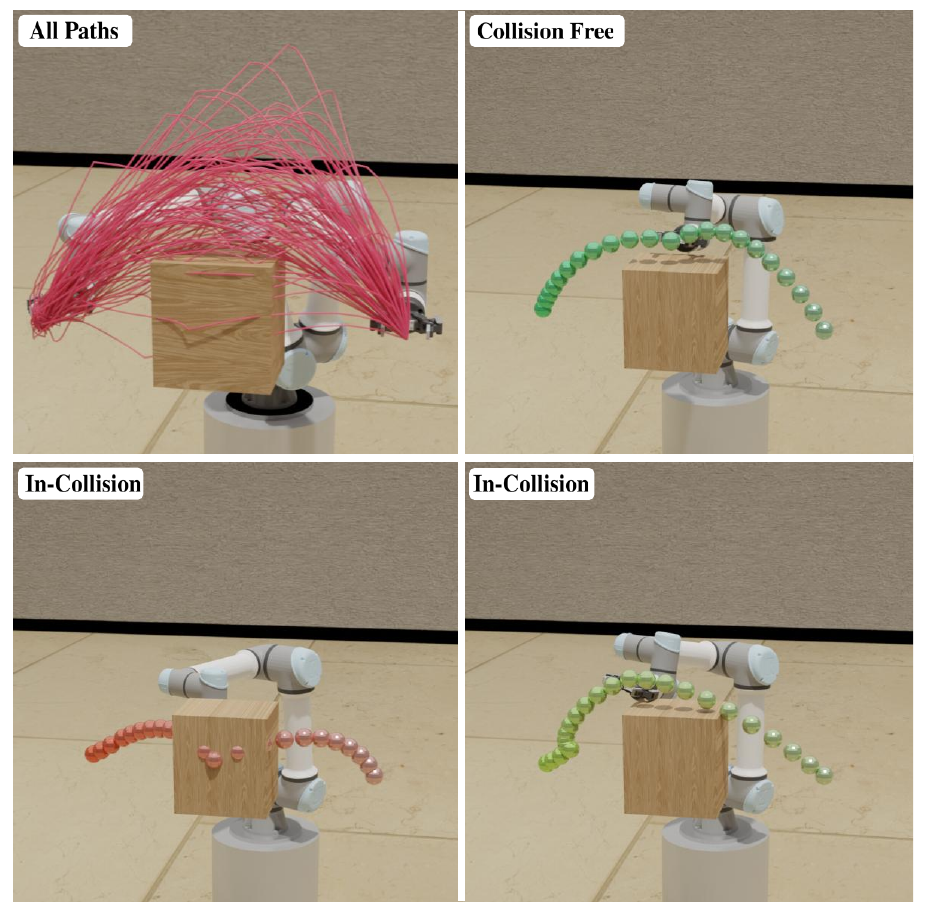}
    \centering
    \captionof{figure}{\small \textbf{Best-of-$N$ Sampling.} Qualitative path diversity of \name for a fixed start-goal-scene query. \name generates diverse path proposals which enables best-of-$N$ sampling to select collision-free candidates after collision checking.}
    \label{fig:multimodal}
\end{figure}

Additionally, Table~\ref{tab:multimodality} and Figure~\ref{fig:multimodal} provide further insight into why best-of-$N$ sampling is effective for \name. For a fixed start-goal-scene query, the flow policy generates diverse path proposals in both joint and end-effector space, with many being collision-free. This suggests that \name learns a useful path distribution from which best-of-$N$ can select collision-free paths.

\begin{table}[htbp]
\centering
\scriptsize
\captionof{table}{\small \textbf{Best-of-$\mathbf{N}$ Sampling.} path-level diversity of \name for a fixed start-goal-scene query. Diversity is measured as the mean pairwise per-waypoint distance across $N$ sampled paths.}
\label{tab:multimodality}
\setlength{\tabcolsep}{3pt}
\resizebox{0.49\textwidth}{!}{%
\begin{tabular}{ccccc}
\toprule
 $N$& \textbf{Joint Diversity (rad)} & \textbf{EE Diversity (m)} & \textbf{Goal-Reaching} & \textbf{Collision-Free} \\
\midrule
100 & $0.092 {\pm} 0.024$& $0.041 {\pm} 0.012$ & 100/100 & 45/100 \\
\bottomrule
\end{tabular}}
\end{table}

\begin{figure}[htbp]
    \includegraphics[width=0.9\linewidth]{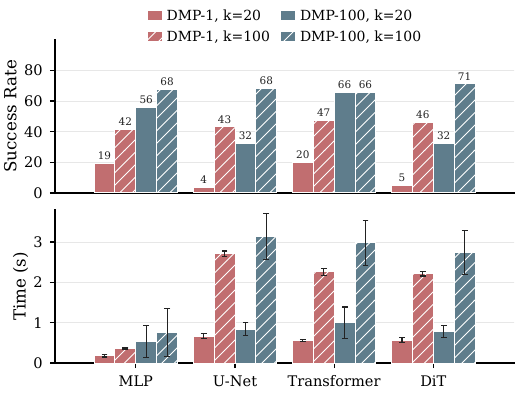}
    \centering
    \captionof{figure}{\small \textbf{Number of Function Evaluation (NFE).} Average planning success rate and planning time of Diffusion Motion Policy with various head architectures and denoising steps (NFE) across all held-out evaluation tasks. ``DMP-1'' and ``DMP-100'' are Diffusion Motion Policy without inference-time optimization ($N=1$) and inference-time optimization ($N=100$), respectively. ``DiT'' denotes DiT-Block Policy~\cite{dasari2025ingredients}.}
    \label{fig:diffusion_timestep}
\end{figure}

\vspace{0.2cm}
\noindent
\textbf{Ablation study: policy head architecture.} We evaluate the effect of policy head design by replacing the proposed MLP-based flow head with U-Net~\cite{chi2025diffusion}, transformer~\cite{chi2025diffusion}, and DiT-Block~\cite{dasari2025ingredients} variants. This ablation tests whether the gains of \name come from the flow-based formulation rather than a specific head architecture. In addition to architecture, we ablate the generative formulation by replacing the flow-based head of \name with a diffusion-based policy head~\cite{chi2025diffusion}, while keeping the rest of the framework unchanged. This baseline uses the same backbone but follows the standard diffusion formulation~\cite{ho2020denoising} with 100 denoising steps during training and inference.

\begin{figure}
    \includegraphics[width=1.0\linewidth]{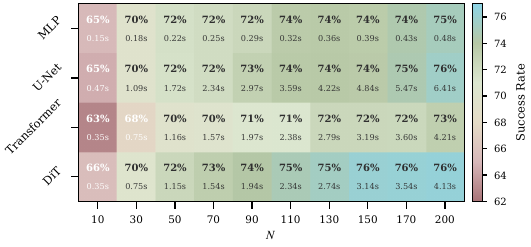}
    \centering
    \captionof{figure}{\small \textbf{Time-to-Solution} for FMP across different policy head architectures with various numbers of sampled paths ($N \in [10, 200]$) at inference.  ``DiT'' denotes DiT-Block Policy~\cite{dasari2025ingredients}.}
    \label{fig:tts}
\end{figure}

As shown in Table~\ref{tab:head_ablation} and Figures~\ref{fig:ablation},~\ref{fig:tts}, \name with the MLP head achieves the fastest planning time among all policy head architectures. The diffusion baseline is slower because it requires 100 denoising steps at inference. While more complex heads can slightly improve success rate on some tasks, they increase model size and architectural complexity, as reported in Table~\ref{tab:ablate-head}.

\begin{figure}
    \includegraphics[width=0.9\linewidth]{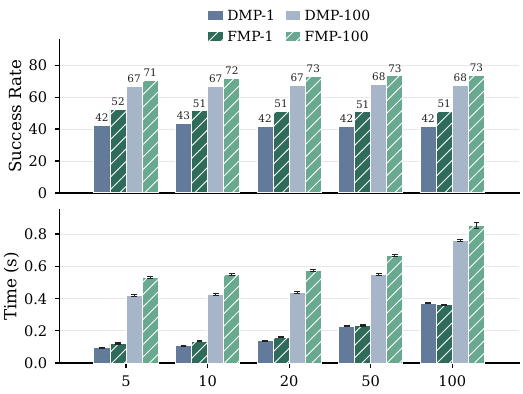}
    \centering
    \captionof{figure}{\small \textbf{Number of Function Evaluation (NFE).} Average planning success and planning time for Diffusion Motion Policy and \name a range of inference steps (NFE) across all held-out planning tasks.  ``DMP-1'' and ``DMP-100'' are Diffusion Motion Policy without inference-time optimization ($N=1$) and inference-time optimization ($N=100$), respectively. ``FMP-1'' and ``FMP-100'' are \name without inference-time optimization ($N=1$) and inference-time optimization ($N=100$), respectively.}
    \label{fig:nfe}
\end{figure}

\begin{figure*}[htbp]
    \includegraphics[width=0.9\linewidth]{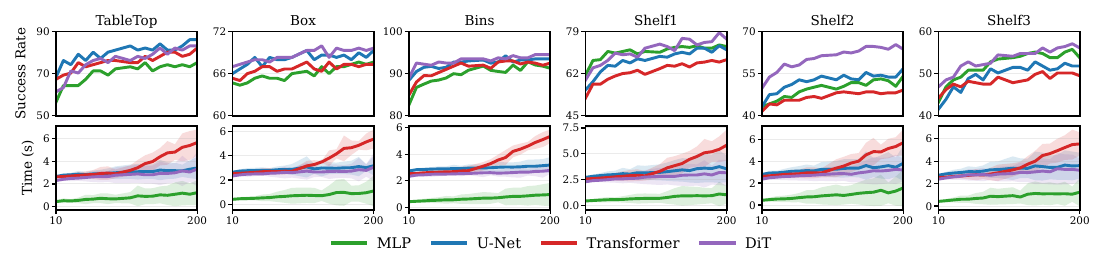}
    \centering
    \captionof{figure}{\small \textbf{Diffusion Motion Policy inference Time Optimization}. Success rate and planning time of Diffusion Motion Policy with various head architectures across a range of trajectories ($N \in [10, 200]$) in inference time optimization. ``DiT'' denotes DiT-Block Policy~\cite{dasari2025ingredients}.}
    \label{fig:ablation_optimization_diffusion}
\end{figure*}

\begin{figure*}[t]
    \includegraphics[width=0.9\linewidth]{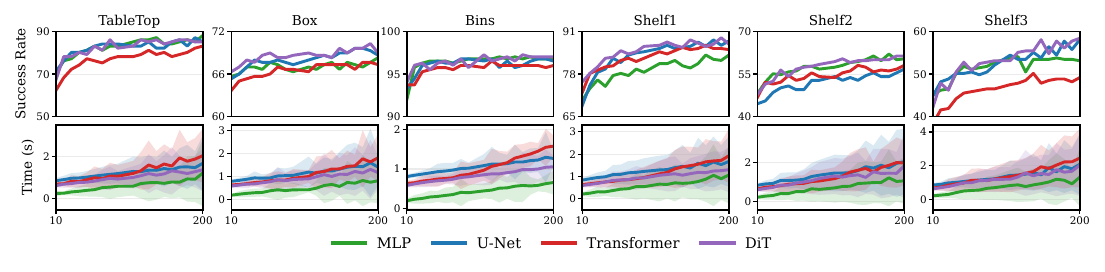}
    \centering
    \captionof{figure}{\small \textbf{\name inference Time Optimization}. Success rate and planning time of \name with various head architectures across a range of trajectories ($N \in [10, 200]$) in inference time optimization. ``DiT'' denotes DiT-Block Policy~\cite{dasari2025ingredients}.}
    \label{fig:ablation_optimization_flow}
\end{figure*}

\begin{figure*}[htbp]
    \includegraphics[width=0.9\linewidth]{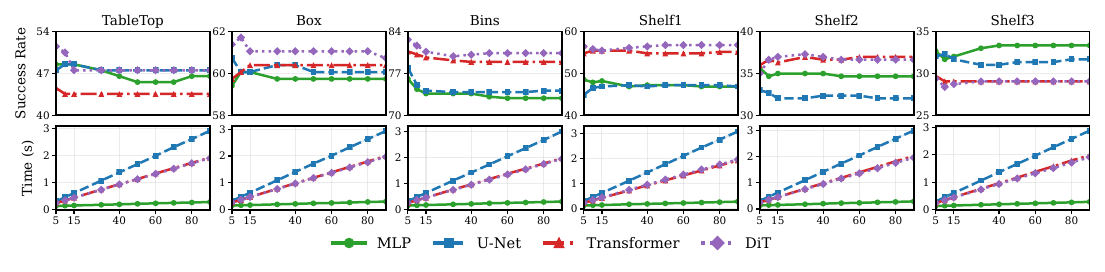}
    \centering
    \captionof{figure}{\small \textbf{Inference Flow Steps.} Success rate and average planning time of \name with various head architectures across a range of inference flow steps ($\text{step} \in [5, 90]$) without inference time optimization ($N=1$). ``DiT'' denotes DiT-Block Policy~\cite{dasari2025ingredients}.}
    \label{fig:euler_steps1}
\end{figure*}

\begin{figure}
    \includegraphics[width=0.9\linewidth]{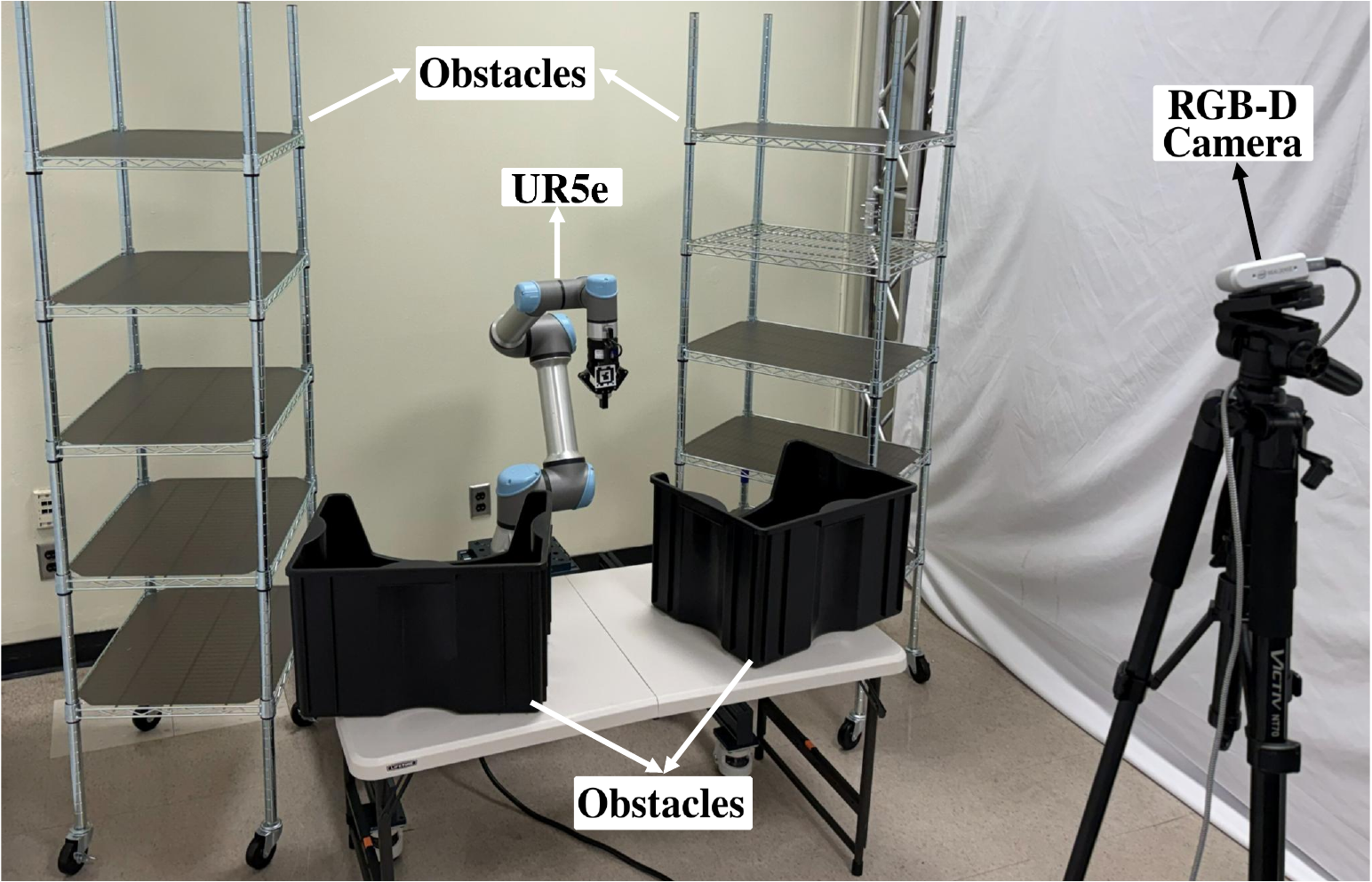}
    \centering
    \captionof{figure}{\small \textbf{Real-World Deployment Setup:} Illustration of the deployment of \name in a real-world environment. A calibrated camera (Intel RealSense D435i RGB-D) is used to capture the workspace point cloud.}
    \label{fig:exp_setup}
\end{figure}

\begin{figure*}[htbp]
    \includegraphics[width=0.9\linewidth]{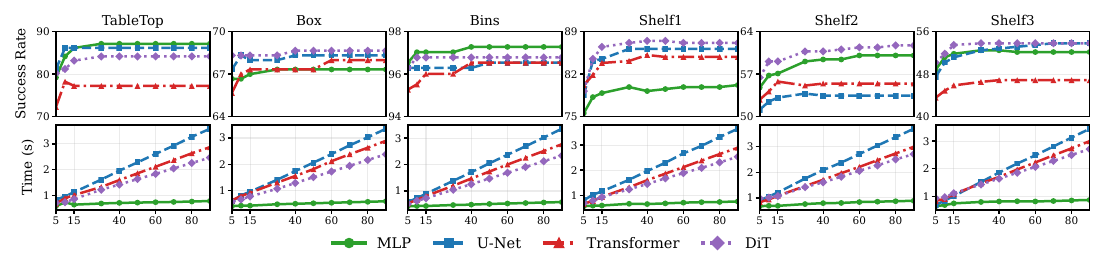}
    \centering
    \captionof{figure}{\small \textbf{Inference Flow Steps.} Success rate and average planning time of \name with various head architectures across a range of inference flow steps ($\text{step} \in [5, 90]$) with inference time optimization ($N=100$). ``DiT'' denotes DiT-Block Policy~\cite{dasari2025ingredients}.}
    \label{fig:euler_steps100}
\end{figure*}

\begin{figure*}[t]
    \includegraphics[width=0.8\linewidth]{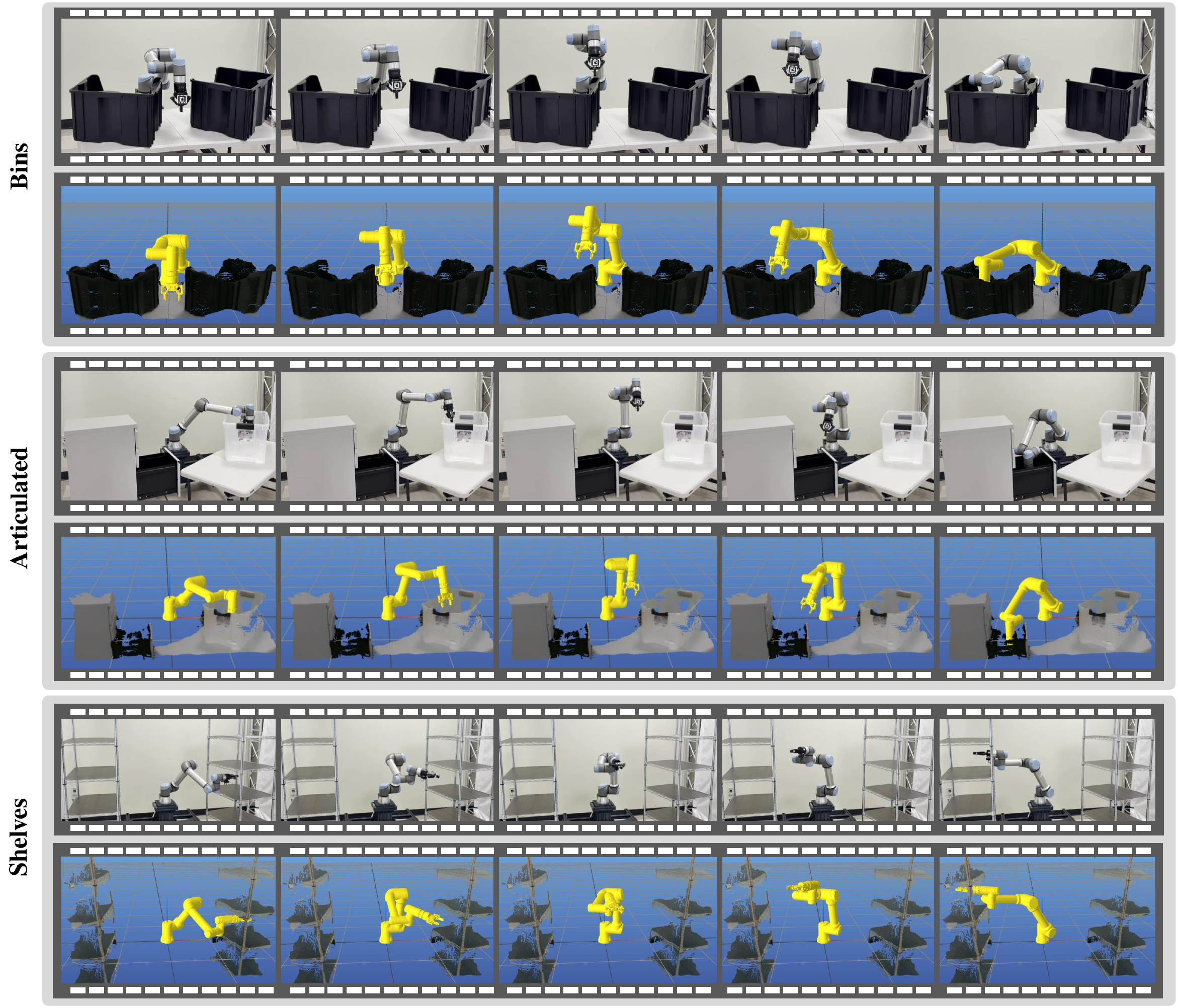}
    \centering
    \captionof{figure}{\small \textbf{\name Deployment in Real-world.} The path for the given motion planning problem is visualized for each environment, illustrating the planned motion execution across subsequent frames.}
    \label{fig:realdemo}
\end{figure*}

We study the effect of inference-time number of function evaluations (NFEs), which corresponds to denoising steps for DMP and flow integration steps for FMP, as shown in Figures~\ref{fig:diffusion_timestep} and~\ref{fig:nfe}. Reducing NFE lowers planning time because fewer model evaluations are used, but it can reduce success rate. Overall, FMP achieves a better success-time trade-off than DMP, reaching higher success with fewer inference steps.

\vspace{0.2cm}
\noindent
\textbf{Ablation study: best-of-$N$ sampling.} We evaluate best-of-$N$ sampling by varying the number of sampled trajectories for \name and Diffusion Motion Policy across different policy heads (Figures~\ref{fig:ablation_optimization_diffusion} and~\ref{fig:ablation_optimization_flow}). Increasing the number of samples improves success rate across all architectures, but also increases planning time because more candidate trajectories must be generated and evaluated. The results also show a trade-off between policy head complexity and inference time. More complex heads can improve the success rate on some tasks, but each trajectory is more expensive to generate. Therefore, planning time grows faster with the number of sampled trajectories for larger policy heads.

\vspace{0.2cm}
\noindent
\textbf{Ablation study: inference flow steps.} We study the effect of the number of Euler solver steps during \name inference across policy head architectures (Figures~\ref{fig:euler_steps1} and~\ref{fig:euler_steps100}). Increasing the number of solver steps, with or without best-of-$N$ sampling, provides little improvement in success rate but increases planning time. This suggests that the policy heads already generate accurate samples with few Euler steps, and additional steps mainly add computational cost.

\vspace{0.2cm}
\noindent
\textbf{Real-world deployment.} We evaluate \name on a UR5e robotic manipulator in real-world environments. One calibrated Intel RealSense D435i RGB-D camera is used to capture point cloud data of the workspace as depicted in Figure~\ref{fig:exp_setup}. AprilTag markers were used to calibrate the camera and estimate the rigid transformation between the camera frame and the robot base frame.

\begin{table}[htbp]
\centering
\captionof{table}{\small \textbf{\name Deployment in Real-world.} Planning success rate of \name performance in real-world evaluation tasks. ``\textbf{FMP-1}'' and ``\textbf{FMP-100}'' are \name without inference-time optimization ($N=1$) and with inference-time optimization ($N=100$), respectively.}
\label{tab: experiments}
\renewcommand{\arraystretch}{1.1}
\begin{tabular}{@{}p{0.20\linewidth}p{0.10\linewidth}p{0.20\linewidth}p{0.10\linewidth}p{0.20\linewidth}p{0.20\linewidth}@{}}
\toprule
\multicolumn{1}{l}{\phantom{Var.}}&\multicolumn{1}{c}{\textbf{Bins}}&\multicolumn{1}{c}{\textbf{Articulated}}&\multicolumn{1}{c}{\textbf{Shelves}}&\multicolumn{1}{c}{\textbf{Total \#}}&\multicolumn{1}{c}{\textbf{Total [\%]}} \\
\toprule
\multicolumn{1}{l}{\textbf{FMP-1}}&\multicolumn{1}{c}{5/10}&\multicolumn{1}{c}{3/10}&\multicolumn{1}{c}{2/10}&\multicolumn{1}{c}{10/30}&\multicolumn{1}{c}{33.4\%} \\
\toprule
\multicolumn{1}{l}{\textbf{FMP-100}}&\multicolumn{1}{c}{\textbf{10/10}}&\multicolumn{1}{c}{\textbf{10/10}}&\multicolumn{1}{c}{\textbf{6/10}}&\multicolumn{1}{c}{\textbf{26/30}}&\multicolumn{1}{c}{\textbf{86.7}\%} \\
\bottomrule
\end{tabular}
\end{table}

The real-world tasks are designed to replicate held-out evaluation scenarios. Table~\ref{tab: experiments} summarizes the planning performance in real-world deployment, and Figure~\ref{fig:realdemo} presents a representative example for each task. 

Across real-world tasks, \name with best-of-$N$ sampling substantially improves success over the base policy ($86.7\%$ vs. $33.4\%$). It solves all bins and articulated tasks, while the base policy achieves only $50\%$ and $30\%$, respectively. Both methods perform worse in shelf environments, likely due to limited similar examples in the training data. Although Best-of-$N$ can improve robustness by selecting among multiple candidates, it remains limited when the base policy does not generate sufficiently diverse feasible trajectories.

\section{Conclusions} \label{sec: conclusion}
In this paper, we introduced \name, an open-loop, end-to-end motion planner that leverages the stochastic formulation of flow matching to enable efficient inference-time best-of-$N$ path sampling. A transformer-based encoder encodes planning information and conditions a denoising flow and is trained via supervised learning on oracle-generated optimal paths. At inference time, \name samples a batch of candidate paths and selects the first collision-free solution, resulting in improved planning success while maintaining planning time. Extensive evaluations demonstrate that \name consistently outperforms both benchmark sampling-based, optimization-based and neural motion planners, with and without inference-time optimization, across held-out planning tasks.

However, several limitations remain. The current formulation assumes static environments and does not explicitly account for dynamic obstacles. Extending \name with reactive controllers, such as Riemannian Motion Policies~\cite{ratliff2018riemannian} or Geometric Fabrics~\cite{van2022geometric}, could enable motion planning in dynamic environments. Additionally, the generated paths may lack smoothness or local optimality and therefore require post-processing before execution on real robots.
\bibliographystyle{IEEEtran}
\bibliography{ref}

@article{soleymanzadeh2026towards,
  title={Towards Generalist Neural Motion Planners for Robotic Manipulators: Challenges and Opportunities},
  author={Soleymanzadeh, Davood and Lopez-Sanchez, Ivan and Su, Hao and Li, Yunzhu and Liang, Xiao and Zheng, Minghui},
  journal={IEEE Transactions on Automation Science and Engineering},
  year={2026},
  publisher={IEEE}
}

@article{noroozi2023conventional,
  title={Conventional, heuristic and learning-based robot motion planning: Reviewing frameworks of current practical significance},
  author={Noroozi, Fatemeh and Daneshmand, Morteza and Fiorini, Paolo},
  journal={Machines},
  volume={11},
  number={7},
  pages={722},
  year={2023},
  publisher={MDPI}
}

@article{lavalle1998rapidly,
  title={Rapidly-exploring random trees: A new tool for path planning},
  author={LaValle, Steven},
  journal={Research Report 9811},
  year={1998},
  publisher={Department of Computer Science, Iowa State University}
}

@article{zucker2013chomp,
  title={Chomp: Covariant hamiltonian optimization for motion planning},
  author={Zucker, Matt and Ratliff, Nathan and Dragan, Anca D and Pivtoraiko, Mihail and Klingensmith, Matthew and Dellin, Christopher M and Bagnell, J Andrew and Srinivasa, Siddhartha S},
  journal={The International journal of robotics research},
  volume={32},
  number={9-10},
  pages={1164--1193},
  year={2013},
  publisher={SAGE Publications Sage UK: London, England}
}

@article{lavalle2001rapidly,
  title={Rapidly-exploring random trees: Progress and prospects: Steven m. lavalle, iowa state university, a james j. kuffner, jr., university of tokyo, tokyo, japan},
  author={LaValle, Steven M and Kuffner, James J},
  journal={Algorithmic and computational robotics},
  pages={303--307},
  year={2001},
  publisher={AK Peters/CRC Press}
}

@article{karaman2011sampling,
  title={Sampling-based algorithms for optimal motion planning},
  author={Karaman, Sertac and Frazzoli, Emilio},
  journal={The international journal of robotics research},
  volume={30},
  number={7},
  pages={846--894},
  year={2011},
  publisher={Sage Publications Sage UK: London, England}
}

@article{das2020forward,
  title={Forward kinematics kernel for improved proxy collision checking},
  author={Das, Nikhil and Yip, Michael C},
  journal={IEEE Robotics and Automation Letters},
  volume={5},
  number={2},
  pages={2349--2356},
  year={2020},
  publisher={IEEE}
}

@article{huang2025prrtc,
  title={prrtc: Gpu-parallel rrt-connect for fast, consistent, and low-cost motion planning},
  author={Huang, Chih H and Jadhav, Pranav and Plancher, Brian and Kingston, Zachary},
  journal={arXiv preprint arXiv:2503.06757},
  year={2025}
}

@inproceedings{thomason2024motions,
  title={Motions in microseconds via vectorized sampling-based planning},
  author={Thomason, Wil and Kingston, Zachary and Kavraki, Lydia E},
  booktitle={2024 IEEE international conference on robotics and automation (ICRA)},
  pages={8749--8756},
  year={2024},
  organization={IEEE}
}

@article{gammell2020batch,
  title={Batch informed trees (bit*): Informed asymptotically optimal anytime search},
  author={Gammell, Jonathan D and Barfoot, Timothy D and Srinivasa, Siddhartha S},
  journal={The International Journal of Robotics Research},
  volume={39},
  number={5},
  pages={543--567},
  year={2020},
  publisher={SAGE Publications Sage UK: London, England}
}

@inproceedings{kalakrishnan2011stomp,
  title={STOMP: Stochastic trajectory optimization for motion planning},
  author={Kalakrishnan, Mrinal and Chitta, Sachin and Theodorou, Evangelos and Pastor, Peter and Schaal, Stefan},
  booktitle={IEEE international conference on robotics and automation},
  pages={4569--4574},
  year={2011},
}

@article{soleymanzadeh2025perfact,
  title={PerFACT: Motion Policy with LLM-Powered Dataset Synthesis and Fusion Action-Chunking Transformers},
  author={Soleymanzadeh, Davood and Liang, Xiao and Zheng, Minghui},
  journal={arXiv preprint arXiv:2512.03444},
  year={2025}
}

@article{dalal2024neural,
  title={Neural mp: A generalist neural motion planner},
  author={Dalal, Murtaza and Yang, Jiahui and Mendonca, Russell and Khaky, Youssef and Salakhutdinov, Ruslan and Pathak, Deepak},
  journal={arXiv preprint arXiv:2409.05864},
  year={2024}
}

@article{yang2025deep,
  title={Deep reactive policy: Learning reactive manipulator motion planning for dynamic environments},
  author={Yang, Jiahui and Liu, Jason Jingzhou and Li, Yulong and Khaky, Youssef and Shaw, Kenneth and Pathak, Deepak},
  journal={arXiv preprint arXiv:2509.06953},
  year={2025}
}

@article{qureshi2020motion,
  title={Motion planning networks: Bridging the gap between learning-based and classical motion planners},
  author={Qureshi, Ahmed Hussain and Miao, Yinglong and Simeonov, Anthony and Yip, Michael C},
  journal={IEEE Transactions on Robotics},
  volume={37},
  number={1},
  pages={48--66},
  year={2020},
}

@article{johnson2023learning,
  title={Learning sampling dictionaries for efficient and generalizable robot motion planning with transformers},
  author={Johnson, Jacob J and Qureshi, Ahmed H and Yip, Michael C},
  journal={IEEE Robotics and Automation Letters},
  volume={8},
  number={12},
  pages={7946--7953},
  year={2023},
}

@inproceedings{kim2023pairwisenet,
  title={PairwiseNet: Pairwise collision distance learning for high-dof robot systems},
  author={Kim, Jihwan and Park, Frank C},
  booktitle={Conference on Robot Learning},
  pages={2863--2877},
  year={2023},
  organization={PMLR}
}

@inproceedings{song2024graph,
  title={Graph-based 3D Collision-distance Estimation Network with Probabilistic Graph Rewiring},
  author={Song, Minjae and Kim, Yeseung and Kim, Min Jun and Park, Daehyung},
  booktitle={IEEE International Conference on Robotics and Automation},
  pages={10939--10945},
  year={2024},
}

@article{carvalho2025motion,
  title={Motion planning diffusion: Learning and adapting robot motion planning with diffusion models},
  author={Carvalho, Joao and Le, An T and Kicki, Piotr and Koert, Dorothea and Peters, Jan},
  journal={IEEE Transactions on Robotics},
  year={2025},
  publisher={IEEE}
}

@article{yan2025m,
  title={M 2 diffuser: Diffusion-based trajectory optimization for mobile manipulation in 3d scenes},
  author={Yan, Sixu and Zhang, Zeyu and Han, Muzhi and Wang, Zaijin and Xie, Qi and Li, Zhitian and Li, Zhehan and Liu, Hangxin and Wang, Xinggang and Zhu, Song-Chun},
  journal={IEEE Transactions on Pattern Analysis and Machine Intelligence},
  year={2025},
  publisher={IEEE}
}

@article{soleymanzadeh2025simpnet,
  title={SIMPNet: Spatial-Informed Motion Planning Network},
  author={Soleymanzadeh, Davood and Liang, Xiao and Zheng, Minghui},
  journal={IEEE Robotics and Automation Letters},
  year={2025},
  publisher={IEEE}
}

@article{lipman2022flow,
  title={Flow matching for generative modeling},
  author={Lipman, Yaron and Chen, Ricky TQ and Ben-Hamu, Heli and Nickel, Maximilian and Le, Matt},
  journal={arXiv preprint arXiv:2210.02747},
  year={2022}
}

@inproceedings{strub2020advanced,
  title={Advanced BIT*(ABIT*): Sampling-based planning with advanced graph-search techniques},
  author={Strub, Marlin P and Gammell, Jonathan D},
  booktitle={2020 IEEE International Conference on Robotics and Automation (ICRA)},
  pages={130--136},
  year={2020},
  organization={IEEE}
}

@inproceedings{zhang2024flexible,
  title={Flexible informed trees (FIT*): Adaptive batch-size approach in informed sampling-based path planning},
  author={Zhang, Liding and Bing, Zhenshan and Chen, Kejia and Chen, Lingyun and Cai, Kuanqi and Zhang, Yu and Wu, Fan and Krumbholz, Peter and Yuan, Zhilin and Haddadin, Sami and others},
  booktitle={2024 IEEE/RSJ International Conference on Intelligent Robots and Systems (IROS)},
  pages={3146--3152},
  year={2024},
  organization={IEEE}
}

@article{qureshi2020neural,
  title={Neural manipulation planning on constraint manifolds},
  author={Qureshi, Ahmed H and Dong, Jiangeng and Choe, Austin and Yip, Michael C},
  journal={IEEE Robotics and Automation Letters},
  volume={5},
  number={4},
  pages={6089--6096},
  year={2020},
  publisher={IEEE}
}

@article{mukadam2018continuous,
  title={Continuous-time Gaussian process motion planning via probabilistic inference},
  author={Mukadam, Mustafa and Dong, Jing and Yan, Xinyan and Dellaert, Frank and Boots, Byron},
  journal={The International Journal of Robotics Research},
  volume={37},
  number={11},
  pages={1319--1340},
  year={2018},
  publisher={SAGE Publications Sage UK: London, England}
}

@article{cohn2025non,
  title={Non-Euclidean motion planning with graphs of geodesically convex sets},
  author={Cohn, Thomas and Petersen, Mark and Simchowitz, Max and Tedrake, Russ},
  journal={The International Journal of Robotics Research},
  volume={44},
  number={10-11},
  pages={1840--1862},
  year={2025},
  publisher={Sage Publications Sage UK: London, England}
}

@inproceedings{nguyen2025flowmp,
  title={Flowmp: Learning motion fields for robot planning with conditional flow matching},
  author={Nguyen, Khang and Le, An T and Pham, Tien and Huber, Manfred and Peters, Jan and Vu, Minh Nhat},
  booktitle={2025 IEEE/RSJ International Conference on Intelligent Robots and Systems (IROS)},
  pages={11291--11297},
  year={2025},
  organization={IEEE}
}

@inproceedings{fishman2024avoid,
  title={Avoid everything: Model-free collision avoidance with expert-guided fine-tuning},
  author={Fishman, Adam and Walsman, Aaron and Bhardwaj, Mohak and Yuan, Wentao and Sundaralingam, Balakumar and Boots, Byron and Fox, Dieter},
  booktitle={CoRL Workshop on Safe and Robust Robot Learning for Operation in the Real World},
  year={2024}
}

@article{chi2025diffusion,
  title={Diffusion policy: Visuomotor policy learning via action diffusion},
  author={Chi, Cheng and Xu, Zhenjia and Feng, Siyuan and Cousineau, Eric and Du, Yilun and Burchfiel, Benjamin and Tedrake, Russ and Song, Shuran},
  journal={The International Journal of Robotics Research},
  volume={44},
  number={10-11},
  pages={1684--1704},
  year={2025},
  publisher={Sage Publications Sage UK: London, England}
}

@inproceedings{dasari2025ingredients,
  title={The ingredients for robotic diffusion transformers},
  author={Dasari, Sudeep and Mees, Oier and Zhao, Sebastian and Srirama, Mohan Kumar and Levine, Sergey},
  booktitle={2025 IEEE International Conference on Robotics and Automation (ICRA)},
  pages={15617--15625},
  year={2025},
  organization={IEEE}
}

@article{bjorck2025gr00t,
  title={Gr00t n1: An open foundation model for generalist humanoid robots},
  author={Bjorck, Johan and Casta{\~n}eda, Fernando and Cherniadev, Nikita and Da, Xingye and Ding, Runyu and Fan, Linxi and Fang, Yu and Fox, Dieter and Hu, Fengyuan and Huang, Spencer and others},
  journal={arXiv preprint arXiv:2503.14734},
  year={2025}
}

@inproceedings{fishman2023motion,
  title={Motion policy networks},
  author={Fishman, Adam and Murali, Adithyavairavan and Eppner, Clemens and Peele, Bryan and Boots, Byron and Fox, Dieter},
  booktitle={conference on Robot Learning},
  pages={967--977},
  year={2023},
  organization={PMLR}
}

@article{black2024pi_0,
  title={$\pi_0 $: A Vision-Language-Action Flow Model for General Robot Control},
  author={Black, Kevin and Brown, Noah and Driess, Danny and Esmail, Adnan and Equi, Michael and Finn, Chelsea and Fusai, Niccolo and Groom, Lachy and Hausman, Karol and Ichter, Brian and others},
  journal={arXiv preprint arXiv:2410.24164},
  year={2024}
}

@article{chen2018neural,
  title={Neural ordinary differential equations},
  author={Chen, Ricky TQ and Rubanova, Yulia and Bettencourt, Jesse and Duvenaud, David K},
  journal={Advances in neural information processing systems},
  volume={31},
  year={2018}
}

@article{zhou2024diffusion,
  title={Diffusion model predictive control},
  author={Zhou, Guangyao and Swaminathan, Sivaramakrishnan and Raju, Rajkumar Vasudeva and Guntupalli, J Swaroop and Lehrach, Wolfgang and Ortiz, Joseph and Dedieu, Antoine and L{\'a}zaro-Gredilla, Miguel and Murphy, Kevin},
  journal={arXiv preprint arXiv:2410.05364},
  year={2024}
}

@article{qi2017pointnet++,
  title={Pointnet++: Deep hierarchical feature learning on point sets in a metric space},
  author={Qi, Charles Ruizhongtai and Yi, Li and Su, Hao and Guibas, Leonidas J},
  journal={Advances in neural information processing systems},
  volume={30},
  year={2017}
}

@article{paszke2017automatic,
  title={Automatic differentiation in pytorch},
  author={Paszke, Adam and Gross, Sam and Chintala, Soumith and Chanan, Gregory and Yang, Edward and DeVito, Zachary and Lin, Zeming and Desmaison, Alban and Antiga, Luca and Lerer, Adam},
  year={2017}
}

@inproceedings{sundaralingam2023curobo,
  title={Curobo: Parallelized collision-free robot motion generation},
  author={Sundaralingam, Balakumar and Hari, Siva Kumar Sastry and Fishman, Adam and Garrett, Caelan and Van Wyk, Karl and Blukis, Valts and Millane, Alexander and Oleynikova, Helen and Handa, Ankur and Ramos, Fabio and others},
  booktitle={IEEE International Conference on Robotics and Automation},
  pages={8112--8119},
  year={2023},
}

@inproceedings{kuffner2000rrt,
  title={RRT-connect: An efficient approach to single-query path planning},
  author={Kuffner, James J and LaValle, Steven M},
  booktitle={Proceedings 2000 ICRA. Millennium conference. IEEE international conference on robotics and automation. Symposia proceedings (Cat. No. 00CH37065)},
  volume={2},
  pages={995--1001},
  year={2000},
  organization={IEEE}
}

@article{sucan2012open,
  title={The open motion planning library},
  author={Sucan, Ioan A and Moll, Mark and Kavraki, Lydia E},
  journal={IEEE Robotics \& Automation Magazine},
  volume={19},
  number={4},
  pages={72--82},
  year={2012},
  publisher={IEEE}
}

@article{soleymanzadeh2026gaide,
  title={GAIDE: Graph-based Attention Masking for Spatial-and Embodiment-aware Motion Planning},
  author={Soleymanzadeh, Davood and Liang, Xiao and Zheng, Minghui},
  journal={arXiv preprint arXiv:2603.04463},
  year={2026}
}

@article{ho2020denoising,
  title={Denoising diffusion probabilistic models},
  author={Ho, Jonathan and Jain, Ajay and Abbeel, Pieter},
  journal={Advances in neural information processing systems},
  volume={33},
  pages={6840--6851},
  year={2020}
}

@article{ratliff2018riemannian,
  title={Riemannian motion policies},
  author={Ratliff, Nathan D and Issac, Jan and Kappler, Daniel and Birchfield, Stan and Fox, Dieter},
  journal={arXiv preprint arXiv:1801.02854},
  year={2018}
}

@article{van2022geometric,
  title={Geometric fabrics: Generalizing classical mechanics to capture the physics of behavior},
  author={Van Wyk, Karl and Xie, Mandy and Li, Anqi and Rana, Muhammad Asif and Babich, Buck and Peele, Bryan and Wan, Qian and Akinola, Iretiayo and Sundaralingam, Balakumar and Fox, Dieter and others},
  journal={IEEE Robotics and Automation Letters},
  volume={7},
  number={2},
  pages={3202--3209},
  year={2022},
  publisher={IEEE}
}

\end{document}